\documentclass[10pt,twocolumn,letterpaper]{article}

\usepackage[arxiv]{cvpr}      % To produce the REVIEW version
\usepackage[dvipsnames]{xcolor}

\definecolor{cvprblue}{rgb}{0.21,0.49,0.74}
\usepackage[pagebackref,breaklinks,colorlinks,citecolor=cvprblue]{hyperref}
\usepackage{makecell}
\usepackage{tabularx}
\usepackage{multicol}

\def\paperID{3002} % *** Enter the Paper ID here
\def\confName{CVPR}
\def\confYear{2024}

\title{Causal-CoG: A Causal-Effect Look at Context Generation for Boosting Multi-modal Language Models}
\author{Shitian Zhao\textsuperscript{1} 
\quad Zhuowan Li\textsuperscript{2} 
\quad Yadong Lu\textsuperscript{1} 
\quad Alan Yuille\textsuperscript{2} 
\quad Yan Wang\textsuperscript{1} 
\\
\textsuperscript{1}Shanghai Key Laboratory of Multidimensional Information Processing, East China Normal University
\\ 
\textsuperscript{2}Johns Hopkins University}
\begin{document}
 \maketitle
 \begin{abstract}
While Multi-modal Language Models (\textit{MLMs}) demonstrate impressive multimodal ability, they still struggle on providing factual and precise responses for tasks like visual question answering (\textit{VQA}). 
In this paper, we address this challenge from the perspective of contextual information. We propose Causal Context Generation, \textbf{Causal-CoG}, which is a prompting strategy that engages contextual information to enhance precise VQA during inference. Specifically, we prompt MLMs to generate contexts, i.e, text description of an image, and engage the generated contexts for question answering. Moreover, we investigate the advantage of contexts on VQA from a causality perspective, introducing causality filtering to select samples for which contextual information is helpful. To show the effectiveness of Causal-CoG, we run extensive experiments on 10 multimodal benchmarks and show consistent improvements, \emph{e.g.}, +6.30\% on POPE, +13.69\% on Vizwiz and +6.43\% on VQAv2 compared to direct decoding, surpassing existing methods. We hope Casual-CoG inspires explorations of context knowledge in multimodal models, and serves as a plug-and-play strategy for MLM decoding. %Besides, {we show causality filter works well on selecting samples on which contextual information is helpful and individual level \textbf{Total Indirect Effect (TIE)} is a better metric in a sample-rerank process.}

%, including MME, SEEDBench, MMBench, POPE, VSR, Winogound, OKVQA, VQAv2, Vizwiz and GQA, showing the effectiveness of Causal-CoG. Besides, 

%First, MLM is prompted to sample multiple descriptions of the given image, as the contextual information. Then MLM give the answer based the given image, the generated context and the question. For each sample, we estimate \textbf{Natural Direct Effect (NDE)} and \textbf{Total Indirect Effect (TIE)} of the given image on the output, which are the concepts in causal inference, to decide whether to apply CoG on this sample. For those applied CoG, the final answer is the aggregation of all the candidates, while for those without CoG, the final answer keep the same with the original answer. We conduct extensive experiments on 10 VQA benchmarks, including MME, SEEDBench, MMBench, POPE, VSR, Winogound, OKVQA, VQAv2, Vizwiz and GQA proving the effectiveness of CoG.
\end{abstract}    
 \section{Introduction}
\label{sec:intro}

% \zhuowan{Alan: personally I am not convinced that this problem should be formulated in terms of causality. My understand of causality (from Judea Pearl's work) is that you need to model the underlying causal dependencies between variables which sometimes requires intervention to determine these dependencies -- e.g., by altering the graph structure and doing controlled experiments. In this paper, however, it is unclear whether causality applies. At least without clearly specifying the difference between a direct question and an indirect one. What is the definition of a direct question? Why can't a direct question involve estimating the context (as part of the direct question)? But there is no need, or time, to debate this now. But I want to raise this as an issue about the paper. And I think, at least, there should be a clear discussion of what you consider to be a direct question.}

\begin{figure}[t]
    \centering
    \includegraphics[width=\linewidth]{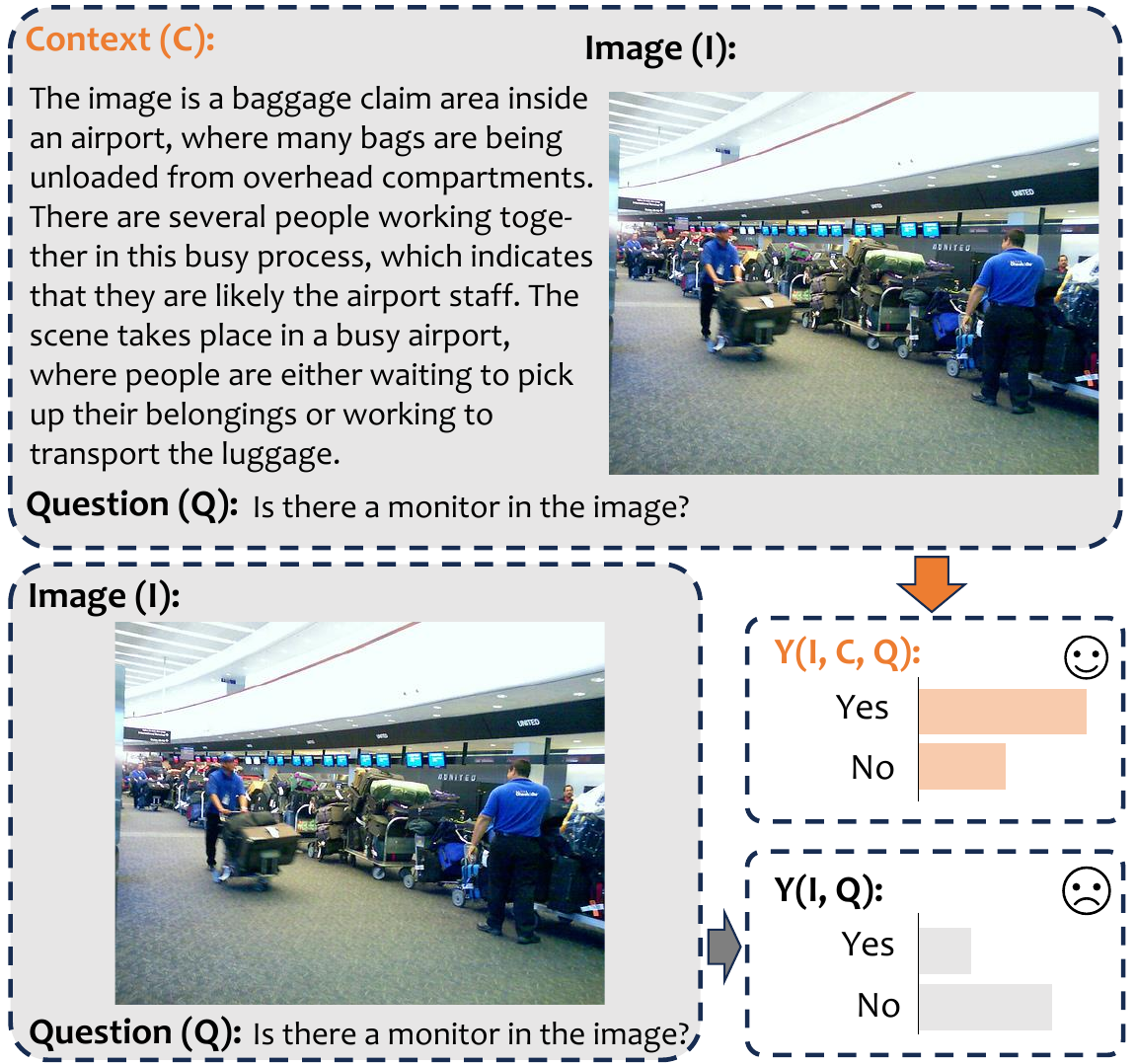}
    \caption{Contextual information helps VQA. In this example, the monitors in the image are too small to see, so the multi-modal language model (MLM) incorrectly predicts ``no'' for the question ``Is there a monitor in this image?''. However, if contextual information is provided, describing that this is an airport baggage hall, MLM can predict the right answer, since there are usually monitors in the baggage hall.}
    % \vspace{-1em}
    \label{context}
\end{figure} 

\begin{figure*}[t]
    \centering
    \includegraphics[width=0.9\linewidth]{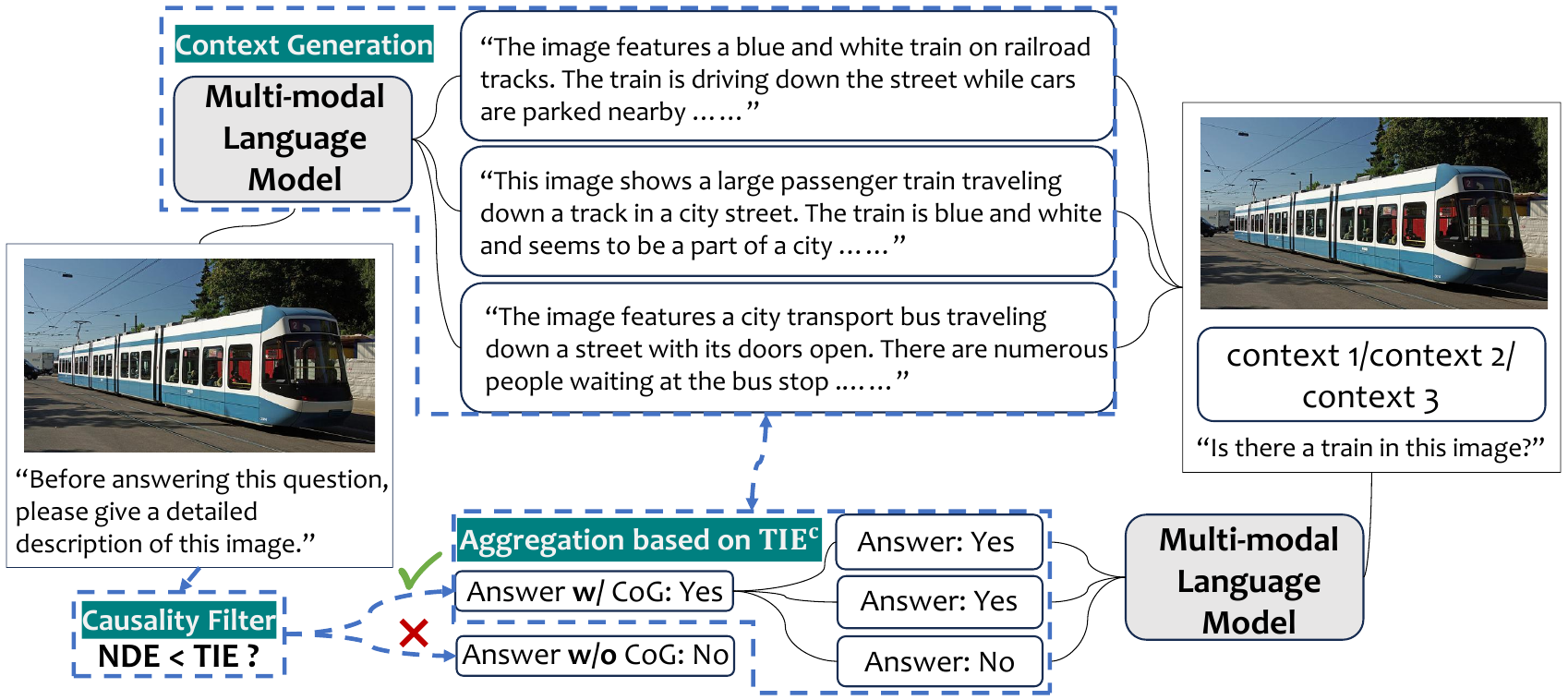}
    \caption{The framework of Causal-CoG: (1) Prompt a multi-modal language model to generate a descriptional context of the image; (2) Repeat this procedure, generating multiple candidates. Calculating and comparing the NDE and TIE of this sample using the generated candidates' answers likelihood distribution, to determine whether to apply CoG on the final answer. (3) For samples determined appropriate for using CoG, aggregate the candidates' answers based on the $\text{TIE}^{\text{c}}$ value of each candidate.}
    % \vspace{-1.0em}
    \label{fig:framework}
    \vspace{-1em}
\end{figure*} 

% Due to the popularity of pretrained autoaggressive  large language model, there are a lot of works that integrade the visual understanding ability into LLMs through adding a pretrained visual encoder and the vision-language alighment module(MLP layer, Q-former) to LLM, such as LLaVA, MiniGPT-4, etc. Due to the powerful LLM in these models and the instruction tuning stage, these models can follow users’ instructions so well and gain good performance on many VQA benchmark, which shows a great potential for building a powerful multi-modality dialog system. However, these models are suffering  the issues of hallucination and weak reasoning ability, the same as LLM. For example, they may give you the wrong answer when you asking the existence of an object in an image and can’t understand the spatial relationship of multiple objects in an image. (a figure) In this paper, we introduce a method called Context Generation (CoG) to boost the VLMs’ visual understanding ability and visual reasoning ability.

Owing to the widespread adoption of Large Language Models (LLM)~\cite{tou23llama,tou23llama2}, there has been a proliferation of research endeavors aimed at integrating visual ability into language models to build Multi-modal Language Models (MLM)~\cite{liu23llava,gao23llamaadapterv2,zhu23minigpt4}. Representative works, \emph{e.g.}, LLaVA~\cite{liu23llava} and MiniGPT-4~\cite{zhu23minigpt4}, incorporates a pretrained visual encoder and a lightweight alignment module to align visual features into LLMs.
Leveraging the intrinsic power of LLM, these models exhibit great capability in following user instructions and show good performance across various multimodal benchmarks, underscoring their potential in multimodal understanding. 

However, similar to LLMs, MLMs grapple with issues related to hallucination\footnote{Hallucination is referred to \textit{``the generated content is nonsensical or unfaithful to the provided source content"}~\cite{ji23hallu} 
%In this paper, hallucination is defined as MLM not able to give factual answers based on the visual content.
}~\cite{Li23POPE}. The models may predict incorrect responses when inquired with misleading queries like the existence of an object in the image, and may struggle to grasp complex relationships among multiple objects in an image~\cite{Liu22vsr}. %\textcolor{red}{here may define hallucination in a more broader way...maybe the following methods cannot solve the hallucination? need details and why they cannot solve}
%Hallucination, \emph{defined as the inability of MLM to provide factual answers based on the visual content in this paper,} 
As the example shown in \cref{context}, when asked the question ``is there a monitor in the image'' for an image containing small monitors that can hardly be seen, current MLM incorrectly predicts the answer as ``no''. 
This inability to provide factual answers based on the visual content is commonly observed in MLMs, due to possible reasons like shortcuts and noise in the training data, lack of modeling capacity for effective alignment of two modalities, etc.

% due to the following three reasons: \textbf{(1)} %Intrinsic flaw of the autoregressive architecture of current language models and training paradigm~\cite{zhang23hallu_llm}. ``Next token prediction" pretraining task forces the model to generate texts autoregressively conditioned on the input and the errors can accumulate. 
% The misalignment between two modalities. Most MLMs use naive alignment modules to align vision and language modalities. For example, LLaVA~\cite{liu23llava} uses a MLP layer to align and MiniGPT4~\cite{zhu23minigpt4} uses a Q-Former~\cite{li23blip2} module to align.
% \textbf{(2)} Noise and error in the multi-modal training data. The training data used during the multi-modal pretraining stage and visual instruction tuning stage contain synthetic data and internet data. Synthetic data is generated by the current powerful LLMs, \emph{e.g.}, ChatGPT~\cite{openai23gpt4,ouyang22instructgpt}, which inherits the hallucination of the generated content. Internet data contains a lot of noise on the internet~\cite{li22blip,radford21clip}.{\textbf{(3)} A relatively small amount of multi-modal training data. Compared to the text data LLM used during pretraining, the amount of multi-modal MLM data used during the alignment training is relatively small. For instance, LLaVA~\cite{liu23llava} use $\sim$150k pairs of multi-modal training data, while LLaMA~\cite{tou23llama}, the LLM backbone of LLaVA, uses 2 trillion tokens to pretrain.} 

There have been explorations boosting the off-the-shelf LLMs in the language community, \emph{e.g.}, chain-of-thought~\cite{kojima22zerocot}, tree-of-thought~\cite{yao23tot} and retrieval-augmented generation~\cite{lewis20rag}, 
% \zhuowan{RLHF (requires training) is not very relevant at a first glance. I can see the point, but drawing this connection will need a large paragraph. So either explain in detail, or remove.}
\emph{etc}. Meanwhile, existing works in the vision-language community improves multimodal models 
%, which can be seen as efforts in eliminating the hallucination issue, 
through shortcut debiasing~\cite{niu21cfvqa} or better alignment of the vision and language modality by various loss functions~\cite{bao22vlmo,li21albef,yu22coca}.
However, these works require training, thus cannot be easily applied to the off-the-shelf MLMs.

In this work, we improve MLM inference from the perspective of contextual knowledge. For example, in \cref{context}, the monitors are too small to see, but if context information like ``this is a baggage claim in airport'' is provided, then the question can be much easier to answer, since a baggage claims usually have monitors on the top. Given this intuition, we take advantage of the context, \ie the description of the image, for more effective question answering. 

% Think about this scenario: you are asked a question about an image. To answer this question, you may first inspect the provided image to gather essential information about it, \emph{e.g.}, what are the objects depicted in the image or what is the style of this image. Then, you are able to give the factual answer. Therefore, the initial inspection step is essential, and the ``context'' collected during this process show its value in answering the question. In fact, several studies, \emph{e.g.}, MM-CoT~\cite{zhang23mmcot} and ScienceQA~\cite{Lu22scienceqa}, have demonstrated the beneficial impact of incorporating contextual information during VQA. In these investigations, context is typically from the dataset itself, or context generation is an integral stage of the training process. Thus, it is inappropriate and inflexible to adopt these methods.

To this end, we propose context generation with a causal-effect look, dubbed as Causal-CoG, which is a prompting technique for MLMs. The method is shown in Fig.~\ref{fig:framework}. Concretely, instead of prompting MLMs to answer questions directly, we first instruct the MLM to generate a description, i.e. context, of the image by employing a simple prompt like ``describe this image'' (rephrased in different flexible ways), then prompt the model to answer the question based on the generated context description. With multiple prompting runs, different context descriptions can be generated, which provides rich information for answering the question. Moreover, to select the most helpful contexts from the multiple generated candidates, we leverage casual inference and take a causal effect look at the contexts. Finally, we propose a candidate aggregation method to attribute greater weight to better candidates considering the impact of the context on the answer. To the best of our knowledge, we are the first to develop ``prompting'' technique for MLMs.

We conduct extensive experiments to show the effectiveness of Casual-CoG. We run experiments on recent standard benchmarks, including MME~\cite{fu2023mme}, SEEDBench~\cite{Li23seedbench}, MMBench~\cite{Liu23mmbench}, POPE~\cite{Li23POPE}, VSR~\cite{Liu22vsr}, and reformed versions of some traditional datasets: VQAv2~\cite{goyal17vqav2}, Vizwiz~\cite{gurari18vizwiz}, GQA~\cite{hudsom19gqa}, OKVQA~\cite{marino19okvqa} and Winoground~\cite{thrush22winoground}, from ReForm-Eval~\cite{li23reformeval}. On all the benchmarks, Casual-CoG leads to consistent improvements, \emph{e.g.}, +6.30\% on POPE, +13.69\% on Vizwiz and +6.43\% on VQAv2, showing the advantage of context knowledge for VQA tasks.
% Our experimental findings unequivocally demonstrate the substantial enhancement in visual understanding and visual reasoning capabilities achieved through the utilization of CoG, with over \red{XXX} improvement. 
To summarize, our contributions are as follows:
\begin{itemize}
    \item We propose Causal-CoG, a training-free decoding strategy that can be easily applied to the off-the-shelf MLMs for generating factual response for VQA. 
    \item Causal-CoG explores the usage of context knowledge, with context filtering and aggregation using causality.
    \item Extensive experiments on 10 datasets show the effectiveness of our method. Causal-CoG consistently boosts the performance of MLMs.
\end{itemize}

 \section{Related Work}
\label{sec:relatedwork}

\noindent
\textbf{Multi-modal Language Models.}
With the remarkable success of Large Language Models (LLM) in the field of Natural Language Processing (NLP), there has been a significant surge of interest in extending LLMs to the multi-modal domains. Noteworthy prior research,such as VisualGPT~\cite{chen22visualgpt} and Frozen~\cite{ts21frozen}, has achieved significant progress in cross-modal tasks. VisualGPT~\cite{chen22visualgpt} employs a series of prompts to facilitate LLMs input, while Frozen~\cite{ts21frozen} trains a visual component to adapt to large language models. Additionally, Flamingo~\cite{jean22flamingo} has demonstrated impressive in-context few-shot learning capabilities by utilizing gated cross-attention to align a pre-trained vision encoder and a language model. Similarly, BLIP2~\cite{li23blip2} leverages a Q-Former to align the visual and language modalities. Building upon these foundations, MiniGPT-4~\cite{zhu23minigpt4} utilizes a limited number of high-quality image-text pairs to align MLM, while InstructBLIP~\cite{dai23instructblip} and LLaVA~\cite{liu23llava} enhance comprehensive MLM performance through instruction tuning. These advancements have significantly contributed to the advancement of MLM.

\noindent
\textbf{Prompt Techniques for LLM.}
Prompt engineering technique has been utilized on LLM to improve its reasoning ability, \emph{e.g.}, CoT~\cite{kojima22zerocot}, ToT~\cite{yao23tot} and GoT~\cite{besta23got}. The strategy of these methods is to first prompt LLM to generate the reasoning path, then based on the reasoning path, LLM can give a more factual answer. LLM may generate multiple reasoning path, and we need to select better reasoning paths or aggregate these paths, \emph{e.g.}, self-consistent~\cite{wang23selfconsistent} and cumulative reasoning~\cite{zhang23cumulativereason}. Prompt engineering technique is under-explored in MLM.

% \zhuowan{Related works can be extended. Can add a paragraph for ``Context in VQA'', citing the papers for VQA debiasing}

\noindent
\textbf{Visual Question Answering with Context.}
Context-aware VQA, which considers contextual information to better understand and answer questions about images, has gained significant attention in recent years. Experiments in ScienceQA~\cite{Lu22scienceqa}, MM-CoT~\cite{zhang23mmcot} and LLaMA-Adapter-V2~\cite{gao23llamaadapterv2} have shown that contextual information can help MLM better answer the visual questions. There are also some works focusing on forcing the model to attend the context when answering questions, \emph{e.g.}, CAD~\cite{shi23cad}.

 \section{Causality in VQA with Context}
\label{sec:preliminaries}

\begin{figure}[t]
    \centering
    \includegraphics[width=0.8\linewidth]{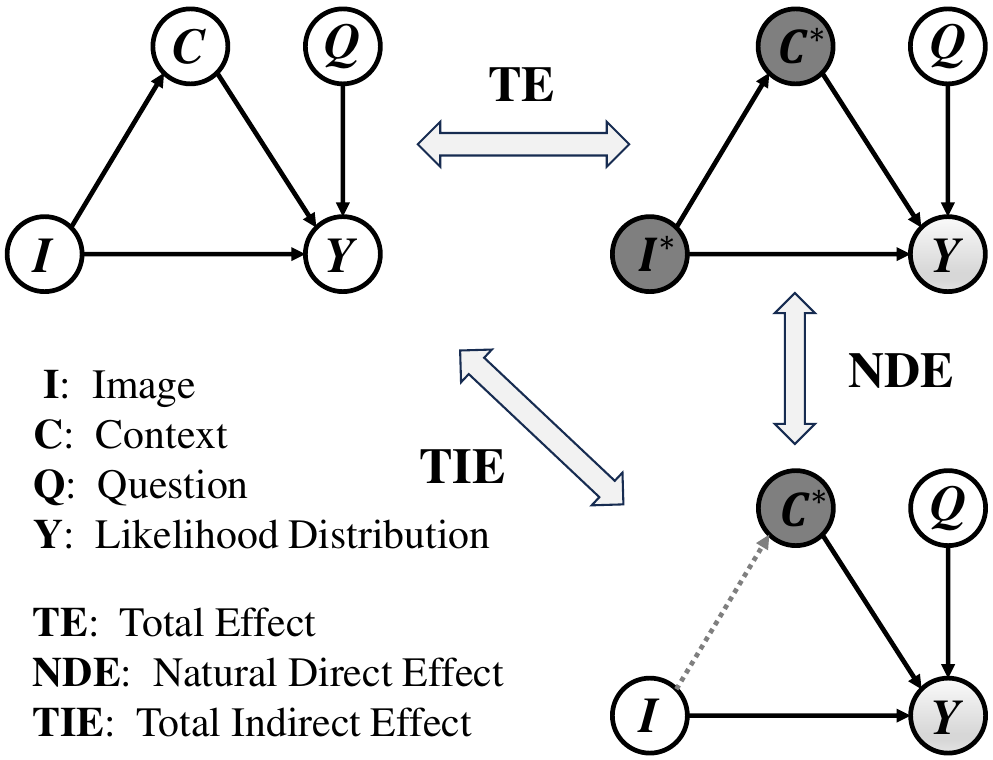}
    \caption{Causal graph of visual question answering with context. Capital letter with ``*" as superscript means not input this variable into MLM when doing VQA.}
    \label{causal}
    \vspace{-1.5em}
\end{figure} 
% In this section, we introduce the basic concepts in causal inference and the basic knowledge that will be used in CoG's causality filter.
In this section, we provide an introduction to fundamental concepts in causal inference, elucidating the foundational knowledge that underpins Causal-CoG's causality filter in Sec.~\ref{sec:causalityfilter}. 
% In this section, we use a capital letter, e.g. C, to stand for a random variable and a capital letter with a subscript character, e.g. $C_{i}$, to stand for the observed value.
\subsection{Causal Graph}
% Causal graph is a directed acyclic graph showing the causal relationship between the nodes, represented as $G = \{V, E\}$. V denoted the set of the variables in this causal graph and E denotes the set of causal-and-effect relationships between two variables. For example, in Figure~\ref{causal}, there exists a causal-and-effect relationship between C and I, so we use a direct line from I to C to represent this relationship.
% In Figure~\ref{causal}, we build the causal graph of VQA with context.

A causal graph is a directed acyclic graph that serves as a graphical representation of the causal relationships between nodes. It is typically denoted as $\mathcal{G} = \{\mathcal{V}, \mathcal{E}\}$, where $\mathcal{V}$ represents the set of variables within the causal graph, and $\mathcal{E}$ denotes the set of causal-effect relationships between pairs of variables. We construct the causal graph specific to VQA with context in Fig.~\ref{causal}. When doing VQA with context, image ($I$), question ($Q$) and context ($C$) are input to MLM, and MLM can output likelihood distribution on different options, termed as $Y$. In Fig.~\ref{causal}, if a causal-effect relationship exists between two variables, \emph{e.g.}, $C$ and $I$, it can be symbolized by $I\xrightarrow{}C$. 
%More details will be provided in Sec.~\ref{sec:causaleffect}.

%In Fig.~\ref{causal}, we construct the causal graph specific to VQA with context.

\subsection{Causal Effect}
\label{sec:causaleffect}
% Causal effect can reflect the comparison of the potential two outcomes with and without a treatment. In Figure~\ref{causal}, I(Image) is a treatment of Y(Logits distribution). There exists two kinds of effect between I and Y: direct effect($I\xrightarrow{}Y$) and indirect effect($I\xrightarrow{}C\xrightarrow{}Y$). In causality's literature, we can calculate the total effect of I on Y through comparing Y(I, C, Q) and Y(Q), which is formulated as:

Causal effect serves as a measure that assesses the contrast between potential outcomes with and without a specific treatment. In Fig.~\ref{causal}, $I$ functions as a treatment for $Y$. There are two distinct types of effects between $I$ and $Y$: the direct effect ($I\xrightarrow{}Y$) and the indirect effect via the generated context ($I\xrightarrow{}C\xrightarrow{}Y$). In the literature of causality, the Total Effect (TE) of $I$ on $Y$ is calculated by comparing $Y(I, C, Q)$ and $Y(Q)$, expressed as:
\begin{equation}
   \text{TE} = \mathbb{E}[Y(I, C, Q) - Y(Q)],
\end{equation}
% TE consists of NDE(Natural Direct Effect) and TIE(Total Indirect Effect). Freezing the variable C, we can calculate NDE by comparing the potential outcome with and without I, which is formulated as:
where $\mathbb{E}[\cdot]$ represents expectation operation and $Y(I, C, Q)$ represents the answer obtained in the VQA task, taking $I$, $C$, and $Q$ as input. The TE encompasses two essential components: the Natural Direct Effect (NDE) and the Total Indirect Effect (TIE). By fixing the variable $C$, the NDE is computed by contrasting the potential outcomes with and without $I$, as formulated by:
\begin{equation}
\label{eq:nde}
    \text{NDE} = \mathbb{E}[Y(I, Q) - Y(Q)].
\end{equation}
% TIE is the differece between TE and NDE:
TIE represents the difference between TE and NDE:
\begin{equation}
\label{eq:tie}
    \text{TIE}=\text{TE}-\text{NDE}=\mathbb{E}[Y(I, C, Q)-Y(I, Q)].
\end{equation}

% The subtracting character in these formulations here actually means the comparison between the two kinds of outcomes, and we use JS-Divergence to instantiate the comparising procidure. Therefore, the practical calculating methods of TE, NDE and TIE are listed here:
% It is important to note that the subtraction in these formulations signifies a comparison between two types of outcomes, and this comparison is instantiated using the Jensen-Shannon Divergence (JSD) in our work. Therefore, the practical calculation methods for TE, NDE, and TIE are as follows:
% \begin{equation}
% \text{TE} = \frac{1}{N}\sum_{i=1}^{N} \text{JSD} (Y_{i}(I_{const},C_{i},Q_{const}), Y_{i}(Q_{const}))
% \end{equation}
% \begin{equation}
% \text{NDE} = \frac{1}{N}\sum_{i=1}^{N} \text{JSD} (Y_{i}(I_{const},Q_{const}), Y_{i}(Q_{const}))
% \end{equation}
% \begin{equation}
% \text{TIE} = \frac{1}{N}\sum_{i=1}^{N} \text{JSD} (Y_{i}(I_{const},C_{i},Q_{const}), Y_{i}(I_{const},Q_{const}))
% \end{equation}

 \section{Causal-CoG}

% \begin{figure}[t]
%     \centering
%     \includegraphics[width=\linewidth]{figures/causalfilter.pdf}
%     \caption{}
%     \label{causal}
% \end{figure} 

% Some exploration has shown that context can boost the visual understanding ability and visual reasoning ability of VLMs (SQA, MM-CoT, LLaMA-Adapter-V2). In these works, the contexts are usually the lectures of the question or the description of the image, which need to be provided by another experts or a retrained model. Different from these methods, we use LLaVA to generate the context by itself without any retraining or finetuning to boost the model’s ability. But during our exploration, we found that the generated contexts are not always helpful for solving the questions: the context may be wrong or provide helpful information. So, we filter the samples through a causality mechanism to decide using the CoG or not. For those using the CoG, we aggregate the results to get the final answer. 
Prior investigations have demonstrated the potential of context in augmenting the visual understanding and visual reasoning capabilities of MLM, \emph{e.g.}, ScienceQA~\cite{Lu22scienceqa}, MM-CoT~\cite{zhang23mmcot}, and LLaMA-Adapter-V2~\cite{gao23llamaadapterv2}. In these prior studies, the contextual information has typically consisted of rationale or image descriptions, often necessitating input from domain experts or the utilization of retrained models. Different from these approaches, we directly employ this Multi-modal Language Model by using a simple prompt to autonomously generate ``context", bypassing the need for retraining the model to generate specific ``context". 

However, our exploration has revealed that the utility of the generated contexts is not consistently beneficial for addressing the questions posed. These contexts may sometimes be inaccurate or fail to provide valuable information. As a response to this challenge, we propose a causality-based filtering mechanism to determine whether to employ CoG\footnote{CoG in this paper refers to the process of generating multiple candidates and aggregating their answers as the final answer, \emph{i.e.}, removing the causality filter's judgment in Causal-CoG.} or not. For those instances where CoG is deemed appropriate, we aggregate the outcomes to derive the final answer.
\vspace{-0.4em}

\subsection{Context Generation}

% There are different types of contexts to help the VQA task, such as the lecture (SQA) and description of the image(LLaMA-Adapter-V2). In CoG, we prompt the model to generate the description of the given image using: “Before answering this question, please give the detailed description of this image.” And we call the generated description “Context”. 
Various forms of contextual information can be employed to facilitate the VQA task, including lecture content, as exemplified in the case of MM-CoT~\cite{zhang23mmcot}, and image descriptions, as utilized in LLaMA-Adapter-V2~\cite{gao23llamaadapterv2}. Within the framework of CoG, we instruct the model to produce a detailed description of the provided image through the use of the following prompt: ``Before answering this question, please give a detailed description of this image." This generated description is subsequently referred to as the ``Context." Note that the MLM which generates the context is the same as the MLM where we apply Causal-CoG.

\subsection{Causality Filter}
\label{sec:causalityfilter}

\begin{figure}[t]
    \centering
    \includegraphics[width=\linewidth]{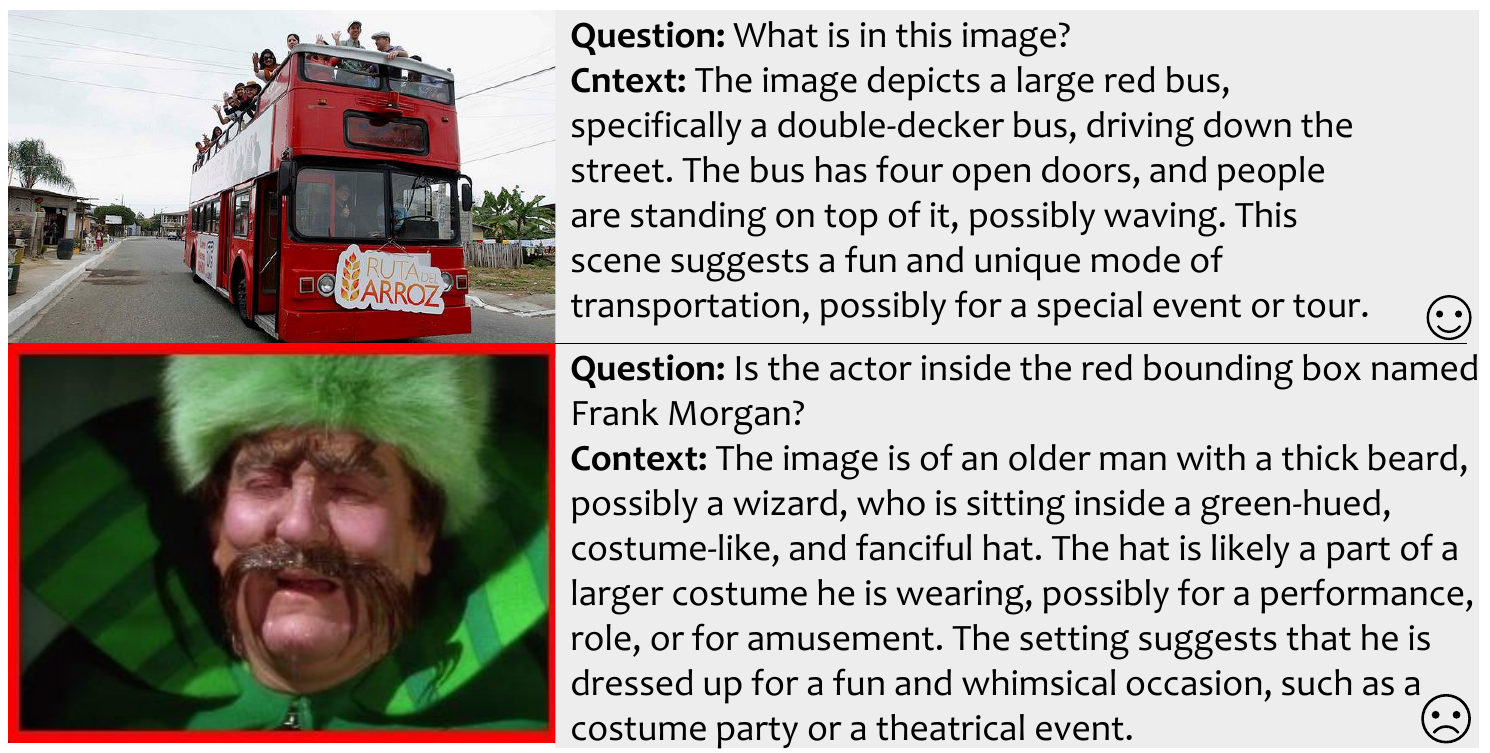}
    \caption{Helpful and unhelpful context. (1) For the upper image, context provides a descriptional information, which is helpful to the question: ``What is it in this image?" (2) For the image below, to determine this actor's name, the common sense of the real world is needed, while context can't provide this kind of information.}
    \vspace{-1.5em}
    \label{fig:helpornot}
\end{figure} 

% As shown in this figure, the generated context may provide the useless even wrong information for answering the question, which can prone the model to give a wrong answer. In this situation, the context can be seen as a noise for the model when answering the question, which can lead to a poorer performance on this question than the model without CoG(figure). So, it is important to consider whether the generated context is helpful on this sample. In other word, we need to design a filter to decide whether the generated context is helpful for the samples.

% Here is the causality graph of the VQA task with the context. We can see that there exist two causal relation.5 between the image and the answer: one is the direct causal path, the other is the indirect causal path. The direct causal path can reflect the direct effect of the given image on the answer, while the indirect effect can reflect the indirect effect, which is through the generated context, of the given image to the answer. In causality’s literature, we can calculate the NDE(natural direct effect) and TIE(total indirect effect) between the given image and the answer. For a sample, if the NDE>TIE, we think that the indirect effect plays a more important role in answering the question, that is, context can help to answer the question. And we choose this sample to apply the CoG technique. In contrast, for those samples with NDE<TIE, we use the answer got by VLM directly as the final answer. 
The generated context provides relevant descriptions of the image. But it is not always helpful in answering the question, as illustrated in Fig.~\ref{fig:helpornot}. This illustration underscores that the generated context has the potential to introduce irrelevant or even erroneous information for answering the question, thereby increasing the likelihood of an incorrect model response. In this scenario, the context can be regarded as a source of noise during the question-answering process, which could result in lower performance compared to a model that operates without CoG, as depicted in Fig.\ref{fig:helpornot}. Therefore, it is important to evaluate whether the generated context is beneficial for a given sample. In essence, we need to devise a filtering mechanism to ascertain the utility of the generated context for individual samples.

The causality graph associated with the VQA task, showed in Fig.~\ref{causal}, reveals the existence of two causal relationships between the image ($I$) and the answers' likelihood distribution ($Y$): one corresponds to the direct causal pathway ($I\xrightarrow{}Y$), while the other represents the indirect causal pathway ($I\xrightarrow{}C\xrightarrow{}Y$). The direct causal path signifies the immediate influence of the provided image on the answer, while the indirect path encapsulates the influence mediated by the generated context. In causality literature, we have the capacity to compute the Natural Direct Effect (NDE) and the Total Indirect Effect (TIE) between the provided $I$ and $Y$.

In Sec.~\ref{sec:preliminaries}, we have introduced how to calculate the NDE and TIE of $I$ on $Y$ (see Eq.~\ref{eq:nde} and Eq.~\ref{eq:tie}). It is important to note that the subtraction in these formulations signifies a comparison between two types of outcomes, and this comparison is instantiated using the Jensen-Shannon Divergence (JSD) in our work. Therefore, the practical calculation methods for NDE, and TIE are as follows:
\begin{align}
&\text{NDE} = \mathbb{E}[\text{JSD}(Y(I,Q), Y(Q))],\\
&\text{TIE} = \mathbb{E}[\text{JSD}(Y(I,C,Q), Y(I,Q))].
\end{align}
Since $\mathbb{E}[\cdot]$ represents the expectation operation, we generate multiple candidates, consisting of generated context and corresponding answers' likelihood distribution, to estimate the expectation value. Overall, $N$ candidates are generated. For the $i$-{th} candidate, we denote the answers' likelihood distribution as $\hat{Y}_{i}$ and the candidate's context as $\hat{C}_{i}$. Here we use a sample with $N$ generated candidates to exemplify the NDE and TIE calculation process:
\begin{align}
    &\text{NDE} = \text{JSD} (\hat{Y}(\hat{I},\hat{Q}), \hat{Y}(\hat{Q})),\\
&\text{TIE} = \frac{1}{N}\sum_{i=1}^{N} \text{JSD} (\hat{Y}_{i}(\hat{I},\hat{C}_{i},\hat{Q}), \hat{Y}_{i}(\hat{I},\hat{Q})),
\vspace{-1em}
\end{align}

where different candidates have different $\hat{Y}_{i}$ and $\hat{C}_{i}$. $I$ and $Q$ remain the same for all the candidates, denoted as $\hat{I}$ and $\hat{Q}$ for this sample. %We use $I_{con}$ and $Q_{con}$ to denote each candidate's I and Q.

{For any given sample, if NDE $<$ TIE, it implies that the indirect effect plays a more pivotal role in responding to the question, signifying that the context can be instrumental in addressing the question effectively. Consequently, we select such samples for the application of the CoG technique. In contrast, for samples whose NDE $>$ TIE, we opt to use the answer generated directly by the MLM as the final answer.}

\subsection{Candidate Aggregation}

% For those samples we choose to apply CoG on, we need to aggregate the candidates’ answers to get the final answer. It is important to notice that though we have filtered out these samples using causality filter, not all the candidates are good ones. As shown in this figure, some generated context may have unconsistent content with the given image. So when aggregating all these candidates’ results, it’s important to consider the context’s quality and effectiveness of each candidate. A trivial strategy is to make the “better” candidates have more conciquence on the final answer. As for how to assign higher weight to the “better” candidates in the aggregation presidure, we take both the similarity between the given image and the generated context and the effect of the context into consideration. 
Given multiple candidates' contexts, we propose an aggregation method to obtain the ultimate response.
%In cases where we decide to employ CoG, the next step involves aggregating the answers provided by various candidates to obtain the ultimate response. 
It's crucial to acknowledge that, despite applying a causality filter to select these samples, not all candidates produce desirable responses. {It's important to assign different weights to different candidates. A trivial approach is to use the answer's likelihood as the weight to aggregate all the candidates' answers or majority voting without weights~\cite{wang23selfconsistent}. But in this work, The determination of higher weight assignment for these candidates hinges on a consideration of the impact of the context on the answer.} 
% Certain generated contexts may exhibit inconsistencies with the provided image. Consequently, when aggregating the results from these candidates, it becomes imperative to assess the quality and effectiveness of each candidate's context.

% One approach involves attributing greater weight to ``better" candidates during the aggregation process. The determination of higher weight assignment for these candidates hinges on a consideration of both the similarity between the given image and the generated context, as well as the impact of the context on the answer. This dual-pronged approach accounts for the quality of the context and its effectiveness in contributing to the final answer.

%\paragraph{Similarity between Image and Context}

% A good candidate’s context should keep consistent with the given image. So, we calculate the similarity between the given image and the generated context, through a pretrained CLIP model,  to score the “consistency” of the image and context, and we denote it as $S_{SIM}$.
% A high-quality candidate's context should maintain a consistent alignment with the provided image. To assess this alignment, we employ a pretrained CLIP model to compute the similarity between the given image and the generated context. This similarity is quantified as $\text{SIM}_i^\text{c}$ for $i$th candidate and is used to gauge the level of ``consistency" between the image and the context.

\noindent
\textbf{Effect of the Context.}
% As shown in this figure, some context may not helps to answer the question, so we think candidates with higher TIE of image on answer via generated context are "better". Similar with Section~\ref{sec:causaleffect}, we calculate the TIE of each candidate denoted as $S_{TIE}$ by:
% \begin{equation}
%     S_{TIE} = jsd(Y(I,C,Q), Y(I,Q))
% \end{equation}
As illustrated in Fig.~\ref{fig:helpornot}, certain contexts may not contribute significantly to the question's answer. Therefore, we consider candidates with a higher TIE value of the image on the answer, mediated through the generated context, to be ``better." Similar to the approach described in Section~\ref{sec:causaleffect}, we compute the individual-level TIE value for $i$th candidate and denote it as $\text{TIE}_i^\text{c}$, which is expressed as:
\vspace{-0.5em}
\begin{equation}
\text{TIE}_i^\text{c} = \text{JSD}(\hat{Y}_i(\hat{I},\hat{C}_i,\hat{Q}), Y(\hat{I},\hat{Q})).
\end{equation}
\vspace{-0.4em}
Here, $\text{TIE}_i^\text{c}$ quantifies the degree of indirect effect and aids in the evaluation of candidate effectiveness in leveraging context to answer questions.

\begin{figure}[t]
    \centering
    \includegraphics[width=0.7\linewidth]{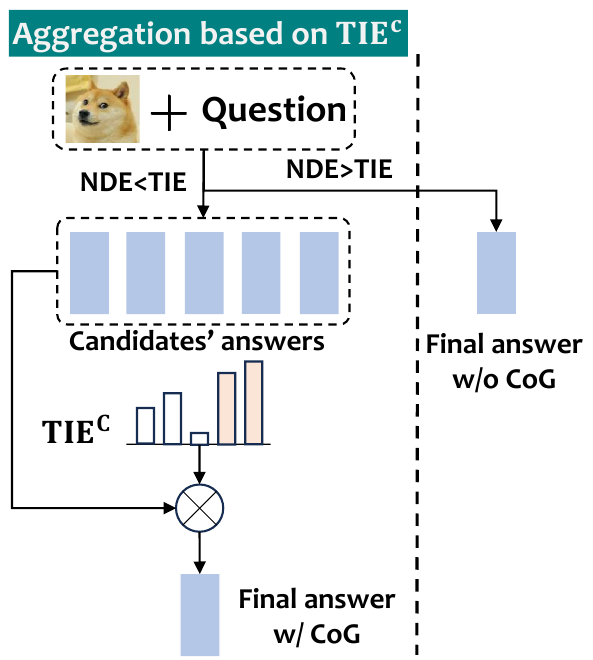}
    \caption{Pipeline of aggregation strategy, exemplified by Top-2 aggregation of 5 candidates' answers. (1) Compare the NDE and TIE values of this sample, determining whether to use CoG to decide the final answer. (2) For samples with TIE\textgreater NDE, aggregate answers based on $\text{TIE}^{\text{c}}$.}
    \label{causalfilter}
    \vspace{-1.3em}
\end{figure} 

\noindent
\textbf{Top-$k$ Aggregation.}
For samples applied with CoG, we need to aggregate the candidates' answers to get the final answer, considering each candidate's $\text{TIE}^\text{c}$. Assuming that we sampled $N$ candidates and each candidate consists of a context and a corresponding answer. After getting the $\text{TIE}^\text{c}$ of each candidate, then
% we first normalize $\text{SIM}_i^\text{c}$ and $\text{TIE}_i^\text{c}$ via:
% \begin{align}
%     &\widetilde{\text{SIM}}_{i}^\text{c} = \text{SIM}_{i}^\text{c}/||\textbf{SIM}^\text{c}||_2,\\
%     &\widetilde{\text{TIE}}_{i}^\text{c} = \text{TIE}_{i}^\text{c}/||\textbf{TIE}^\text{c}||_2,
% \end{align}
\begin{align}
    \textbf{TIE}^\text{c}=[\text{TIE}_{1}^\text{c},...,\text{TIE}_{N}^\text{c}],
\end{align}
% where $\textbf{SIM}^\text{c}=[\text{SIM}_{1}^\text{c},...,\text{SIM}_{N}^\text{c}]$ and $\textbf{TIE}^\text{c}=[\text{TIE}_{1}^\text{c},...,\text{TIE}_{N}^\text{c}]$ indicate the similarity metric and the total indirect effect metric.
where $\textbf{TIE}^\text{c}$ is the set of all candidates' $\text{TIE}^{\text{c}}$.

Next, top-$k$
%\footnote{In this paper, we set k to 5.} 
candidates in ${\textbf{TIE}}^\text{c}$ are picked and whose indices are grouped into a set $\Omega_{\text{TIE}^{\text{c}}}$. Then, we only keep the top-$k$ candidates' values while setting other candidates' values to zero via:
\begin{equation}
    \overline{\text{TIE}}_{i}^\text{c} = \left\{ \begin{array}{cl}
{\text{TIE}}_{i}^\text{c} & : i\in \Omega_\text{SIM} \\
0 & : i\notin  \Omega_\text{SIM}.
\end{array} \right.
\end{equation}
% The same applies to $\widetilde{\text{TIE}}_{i}^\text{c}$ which obtains $\overline{\text{TIE}}_{i}^\text{c}$. Then, the weight of the $i$th candidate is calculated by $w_i^\text{c}=\overline{\text{SIM}}_{i}^\text{c}+\overline{\text{TIE}}_{i}^\text{c}$. We can get the final answer for the sample using the weighted majority vote, according to each candidate's $w_i^\text{c}$.% Given the answer of the $i$th candidate, 
Finally, we can get the final answer for the sample using the weighted majority vote, according to each candidate's $\overline{\text{TIE}}^\text{c}$.

%we keep the top-k candidates' value and set the other candidates' value to zero, acquiring $\beta(\widetilde{\textbf{SIM}}^\text{c})$ and $\beta(\widetilde{\textbf{TIE}}^\text{c})$. Finally, we denote the weight of $i$th candidate as calculated by $w_i=$

%Then we get ranked $S_{SIM}$ list and ranked $S_{TIE}$ list, denoted as $L_{SIM}$ and $L_{TIE}$ respectively. In both of them, we keep the top-k item's value and set the other items' value to zero, denoting the changed list as $L_{SIM}^{'}$ and $L_{TIE}^{'}$. Each candidate get changed $S_{SIM}$ and $S_{TIE}$ according to $L_{SIM}^{'}$ and $L_{TIE}^{'}$. Finally the weight of each candidate is the sum of changed $S_{SIM}$ and $S_{TIE}$, denoting the final weight as $S$. And we can get the final answer using the weighted majority vote, according to each candidate's $S$. Fig.~\ref{causalfilter} illustrate the pipeline of Top-k aggregation.

 \section{Experiment}

\begin{table*}[ht]
\centering
\footnotesize
\begin{tabular}[width=\textwidth]{l|ccccccccccc}
\toprule
                 & \rotatebox{0}{MME} & \rotatebox{0}{\makecell[c]{SEEDBench}} & \rotatebox{0}{\makecell[c]{MMBench}} & \rotatebox{0}{POPE}  & \rotatebox{0}{VSR} &\rotatebox{0}{\makecell[c]{Winoground}*}&\rotatebox{0}{OKVQA*}&\rotatebox{0}{VQAv2*}&\rotatebox{0}{Vizwiz*}&\rotatebox{0}{GQA*}\\
                 \midrule
\makecell[l]{\textit{Ensemble}}  &57.03&39.26&41.43&66.46&54.42&61.25&30.75&42.86&29.47&37.23 \\
\makecell[l]{\textit{One-shot}} &54.72&39.82&41.57&68.66&51.55&62.50&30.56&42.91&32.25&37.55  \\
\midrule
Naive-CoG &60.43&39.20&42.53&68.63&58.10&63.75&30.75&47.81&38.05&39.78 \\
\midrule
LLaVA         &  57.49   &   38.65        &  41.16       &  64.55                  &  55.56   &61.25&30.36&43.10&28.54&37.71\\
\makecell[l]{+Causal-CoG}  &\textbf{61.23}&\textbf{40.66}&\textbf{43.52}&\textbf{70.85}&\textbf{58.92}&\textbf{66.25}&\textbf{32.74}&\textbf{49.44}&\textbf{42.23}&\textbf{41.45}     \\
$\Delta$  &+3.74&+2.01&+2.36&+6.30&+3.36&+5.00&+2.38&+6.43&+13.69&+3.74    \\
\midrule
LLaVA-v1.5      &  \textbf{72.27}   &     45.45      &   43.96      &           85.76  &   58.02    &   63.75&33.92&52.89&39.21&\textbf{49.64}   \\
+Causal-CoG   &71.84&\textbf{45.93}&\textbf{45.43}&\textbf{86.34}&\textbf{61.62}&\textbf{66.25}&\textbf{34.92}&\textbf{54.34}&\textbf{46.64}&48.77 \\
$\Delta$ &-0.43&+0.48&+1.47&+0.64&+3.60&+2.50&+1.00&+1.45&+7.43&-0.87    \\

\bottomrule
\end{tabular}
\caption{Accuracy results on 10 benchmarks. (1) Causal-CoG improves LLaVA~\cite{liu23llava} and LLaVA-v1.5~\cite{liu23llava15}'s performance on 10 benchmarks, \emph{e.g.}, +13.69\% on Vizwiz*~\cite{gurari18vizwiz} and +6.43\% on VQAv2*~\cite{goyal17vqav2}. (2) \textit{Ensemble} and \textit{One-shot} is conducted on LLaVA~\cite{liu23llava}. On some benchmarks, these two methods show a slight improvement. (3) In Naive-CoG, the number of generated candidates is set to 1. Though Naive-CoG's improvement is not as significant as Causal-CoG, it still boosts LLaVA~\cite{liu23llava} on most benchmarks, \emph{e.g.}, +9.51\% on Vizwiz*. (4) \textit{Dataset*} is the split and reformed version of \textit{Dataset} from ReForm-Eval~\cite{li23reformeval}.}
\vspace{-1.5em}
\label{tab:overall}
\end{table*}

\subsection{Setup}
We conduct experiments on 10 VQA benchmarks, comparing the performance of Causal-CoG with competitors. The results show that Causal-CoG can boost the MLM's ability on both perception and reasoning tasks, as well as reduce object hallucination.

\noindent
\textbf{Evaluation Benchmarks.}
We evaluate on 4 comprehensive benchmarks containing multiple subtasks ranging from object identification to visual reasoning: MME~\cite{fu2023mme}, SEEDBench~\cite{Li23seedbench}, MMBench~\cite{Liu23mmbench} and VQAv2~\cite{goyal17vqav2}. Besides, we use POPE~\cite{Li23POPE} for evaluating object hallucination; VSR~\cite{Liu22vsr} and Winoground~\cite{thrush22winoground} for evaluating visual spacial understanding ability; OKVQA~\cite{marino19okvqa} for testing MLM's ability to leverage outside knowledge; Vizwiz~\cite{gurari18vizwiz} for assisting blind people; GQA~\cite{hudsom19gqa} for real-world visual reasoning. Noted that Winoground~\cite{thrush22winoground}, OKVQA~\cite{marino19okvqa}, VQAv2~\cite{goyal17vqav2}, Vizwiz~\cite{gurari18vizwiz} and GQA~\cite{hudsom19gqa} used in this paper are not the original versions, but the split and reformed versions in ReForm-Eval~\cite{li23reformeval} and they are reformed into the single-choice forms, along with $\cdot^{*}$ in the rest of the paper.

% To evaluate perception and reasoning ability of MLM. We split these benchmarks into two sets, listed as follows:
% \begin{itemize}
% \item[-] \textbf{Perception}: POPE, VSR, MME(p), SEEDBench(p) , and MMBench(p)
% \item[-] \textbf{Reasoning}: ScienceQA, MME(r), SEEDBench(r), and MMBench(r)
% \end{itemize}
% where \textit{Benchmark(p/r)} means we take the perception/reasoning subtasks as the test set.

\noindent
\textbf{Multi-modal Language Models.}
We evaluate Causal-CoG on two representative MLMs: LLaVA and LLaVA-v1.5.
\begin{itemize}
    \item[-] \textbf{LLaVA}~\cite{liu23llava}: LLaVA is trained on image-text pair data and visual instruction data. The 7B checkpoint is used in our experiment.
    \item[-] \textbf{LLaVA-v1.5}~\cite{liu23llava15}: LLaVA-v1.5 is an improved version of LLaVA, trained on more multi-modal data. We use the 7B version in our experiment.
    % \item[-] \textbf{MiniGPT4}~\cite{zhu23minigpt4}: MiniGPT4 uses Q-former~\cite{li23blip2} to fuse the visual information. We use its 13B version.
\end{itemize}

\noindent
\textbf{Implementation Details of Candidate Generation.}
We denote the context and the likelihood distribution of the $i$th candidate on all the answer options as $\hat{C}_{i}$ and $\hat{Y}_{i}$.
\begin{itemize}
    \item[-]\textbf{Context}: $\hat{C}_{i}$ is generated by the language model in MLM, thus we can choose the hyperparameters in the sampling strategy to control the generation process of $\hat{C}_{i}$. For all the MLMs mentioned above, we follow the similar setting as \cite{wang23selfconsistent}, setting the temperatur to 0.9~\cite{ari20temp}, and truncating the top-$k$ ($k=40$) tokens with highest probability~\cite{fan18topk}. In our experiments we apply Causal-CoG on LLaVA with 40 candidates and on LLaVA-v1.5 with 20 candidates.
    \item[-]\textbf{Answer's likelihood distribution}: For the $i${th} candidate, the answer options list of this sample is denoted as $L=[A_{1},A_{2},...,A_{M}]$. $\hat{Y}_{i}$ is the likelihood distribution on ${L}$. $A_{j}$'s likelihood, termed as $\hat{Y}_{i,j}$, can be calculated as:
    \begin{equation}
    \begin{aligned}
        \hat{Y}_{i,j}(\hat{X}_{i})=\exp^{\frac{1}{K_{j}}\sum_{k=1}^{K_{j}}}\log P(t_{k}|\hat{X}_{i},t_{1},t_{2},...,t_{k-1}) 
    \end{aligned}
    \end{equation}
    where $\hat{X}_{i}$ is the input MLM take, \emph{e.g.}, $\hat{X}_{i}$ is $[\hat{I},\hat{C}_{i},\hat{Q}]$ in $\hat{Y}_{i}(\hat{I},\hat{C}_{i},\hat{Q})$. $P(t_{k}|\hat{X}_{i},t_{1},t_{2},...,t_{k-1})$ is the probability of generating the $k$th token $t_{k}$ conditioned on $\hat{X}_{i}$ and previous generated tokens $t_{1}\sim t_{k-1}$. $K_{j}$ is the length of the $j${th} answer. As all the benchmarks we use are single-choice formulations, the output answer, regardless of with the application of Causal-CoG or not, is decided based on the options' likelihood distribution, \emph{i.e.}, choose the option with highest likelihood as the output answer.
\end{itemize}

\begin{table*}[ht]
\centering
\footnotesize
\begin{tabular}[width=\textwidth]{l|ccc|ccc|ccc}
\toprule
&\multicolumn{3}{c|}{Cognition}& \multicolumn{3}{c|}{Perception}& \multicolumn{3}{c}{Object hallucination} \\
\midrule
&$\text{MME}_{\text{c}}$&$\text{SEEDBench}_{\text{c}}$&$\text{MMBench}_{\text{c}}$&$\text{MME}_{\text{p}}$&$\text{SEEDBench}_{\text{p}}$&$\text{MMBench}_{\text{p}}$&POPE-P&POPE-A&POPE-R  \\
\midrule
LLaVA   &53.85&37.50&44.30&59.11&40.94&38.02&61.27&57.67&74.71           \\
+Causal-CoG &56.43&39.43&45.26&63.37&43.12&41.78&65.97&63.60&82.99 \\
$\Delta$ &+2.59&+1.93&+0.96&+4.26&+2.18&+3.76&+4.70&+5.93&+8.28\\
\midrule
LLaVA-v1.5 &53.57&46.88&44.37&80.58&42.59&43.55&87.13&82.03&88.11  \\
+Causal-CoG  &53.93&47.17&45.80&79.81&43.45&45.07&86.47&83.57&88.97\\
$\Delta$ &+0.36&+0.29&+1.43&-0.77&+0.86&+1.52&-0.66&+1.54&+0.86 \\

\bottomrule
\end{tabular}
\caption{Results on cognition tasks, perception tasks and object hallucination issues. (1) $\textit{Dataset}_{\text{c}}$ represent the average of results over cognition tasks. While $\textit{Dataset}_{\text{p}}$ represents the average of results over perception tasks. The list of cognition and perception tasks in MME~\cite{fu2023mme}, SEEDBench~\cite{Li23seedbench} and MMBench~\cite{Liu23mmbench} is provided in supplementary material. (2) POPE-P, POPE-A and POPE-R are the abbreviation of POPE-Popular, POPE-Adversarial and POPE-Random respectively, which are three versions of POPE~\cite{Li23POPE}. }
\label{tab:finegrain}
\vspace{-1.0em}
\end{table*}

\noindent
\textbf{Competitors.}
Since the study of MLM's decoding strategy\footnote{MLM's decoding strategy refers to methods can be used directly without retraining or fine-tuning the pretrained MLM during the inference time.} is still under-explored, we compare our Causal-CoG with two methods that can be used directly during the decoding stage of MLM: \textbf{(1) ensemble MLM through various prompts}~\cite{radford21clip} and \textbf{(2) one-shot in-context learning}~\cite{jean22flamingo}. The implementation details are as follows.
\begin{itemize}
    \item[-]\textbf{Ensemble}: We ensemble the outputs by averaging the 5 answers' likelihood distribution using 5 different system prompts\footnote{System prompt refers to the prompt that is contained in every conversation's beginning, telling the MLM to act as an multi-modal chatbot. An example of system prompt is: \textit{``You are a helpful language and vision assistant. And you are able to understand the visual content that the user provides, and assist the user with a variety of tasks using natural language."}~\cite{liu23llava}} on LLaVA. Full list of all system prompts we use is provided in the supplementary material. 
    \item[-]\textbf{One-shot}: Similar to Flamingo~\cite{jean22flamingo}, we use the one-shot in-context learning on LLaVA.
\end{itemize}

\subsection{Main Results}
We apply Causal-CoG on LLaVA and LLaVA-v1.5 respectively, and compare the results with the ones without Causal-CoG. Table~\ref{tab:overall} shows the overall results on 10 benchmarks. Causal-CoG improves the performance of both LLaVA and LLaVA-v1.5 significantly on most of the benchmarks. For LLaVA, Causal-CoG boosts the accuracy by more than 5.00\% over POPE, $\text{Winoground}^{*}$, $\text{VQAv2}^{*}$ and $\text{Vizwiz}^{*}$, 2.00\%-3.00\% absolute improvement over other benchmarks. When it comes to LLaVA-v1.5, we can still observe nearly 2.00\%-3.00\% improvement. It is worth mentioning that LLaVA-v1.5 is an advanced version of LLaVA. Compared with LLaVA, it performs better on all benchmarks significantly, while Causal-CoG can still work on this advanced and more developed MLM, which shows the potential of Causal-CoG for working on well-developed MLMs. In Table~\ref{tab:overall}, \textit{Ensemble} and \textit{One-shot} works on some datasets, improving the accuracy incrementally, while these two methods can cause performance drop on other datesets, \emph{e.g.}, VSR, $\text{VQAv2}^{*}$ and $\text{GQA}^{*}$. 

\noindent
\textbf{Performance on Perception, Recognition and Object Hallucination.}
In MME~\cite{fu2023mme}, SEEDBench~\cite{Li23seedbench} and MMBench~\cite{Liu23mmbench}, all the subtasks are split into two main parts: perception and cognition. Table~\ref{tab:finegrain} shows results on perception tasks and cognition tasks. POPE~\cite{Li23POPE} is a benchmark for testing MLM's object hallucination issue. It contains three versions: POPE-Popular, POPE-Adversarial and POPE-Random. Detailed results of Causal-CoG is included in Table~\ref{tab:finegrain}. We can see that Causal-CoG works better on perception tasks and object hallucination issues than cognition tasks. This may be due to that context MLM generated mainly focuses on the image's description, which is obviously helpful for perception tasks, and may be not helpful for cognition tasks in some cases, \emph{e.g.}, code\_reasoning.

% \begin{table*}[ht]
% \centering
% \footnotesize
% \begin{tabular}[width=\textwidth]{l|ccc|ccc|ccc}
% \toprule
% &\multicolumn{3}{c|}{Cognition}& \multicolumn{3}{c|}{Perception}& \multicolumn{3}{c}{Object hallucination} \\
% \midrule
% &MME(r)&SEEDBench(r)&MMBench(r)&MME(p)&SEEDBench(p)&MMBench(p)&POPE-P&POPE-A&POPE-R  \\
% \midrule
% LLaVA   &53.85&37.50&44.30&59.11&40.94&38.02&61.27&57.67&74.71           \\
% +Causal-CoG &56.43&39.43&45.26&63.37&43.12&41.78&65.97&63.60&82.99 \\
% $\Delta$ &+2.59&+1.93&+0.96&+4.26&+2.18&+3.76&+4.70&+5.93&+8.28\\
% \midrule
% LLaVA-v1.5 &&46.88&44.37&&42.59&43.55&87.13&82.03&88.11  \\
% +Causal-CoG  &&&45.80&&&45.07&86.47&83.57&88.97\\
% $\Delta$ &&&+1.43&&&+1.52&-0.66&+1.54&+0.86 \\

% \bottomrule
% \end{tabular}
% \caption{\textcolor{red}{POPE-P} means xxx...}
% \label{tab:finegrain}
% \end{table*}

% \zhuowan{Show examples for cases where context is helpful, and examples for which context is harmful}

 \subsection{Ablation and Analysis}

% \begin{figure*}[t]
%     \centering
%     \includegraphics[width=0.7\linewidth]{figures/compare.pdf}
%     \caption{}
%     \label{fig:compare}
% \end{figure*} 

% \begin{figure}[t]
%     \centering
%     \includegraphics[width=
% 0.8\linewidth]{figures/acc_vizwiz.pdf}
%     \caption{\textcolor{red}{where is fig 5's text? can put in more analysis}}
%     \label{fig:sim_tie}
% \end{figure} 

\begin{figure}[t]
    \centering
    \includegraphics[width=
\linewidth]{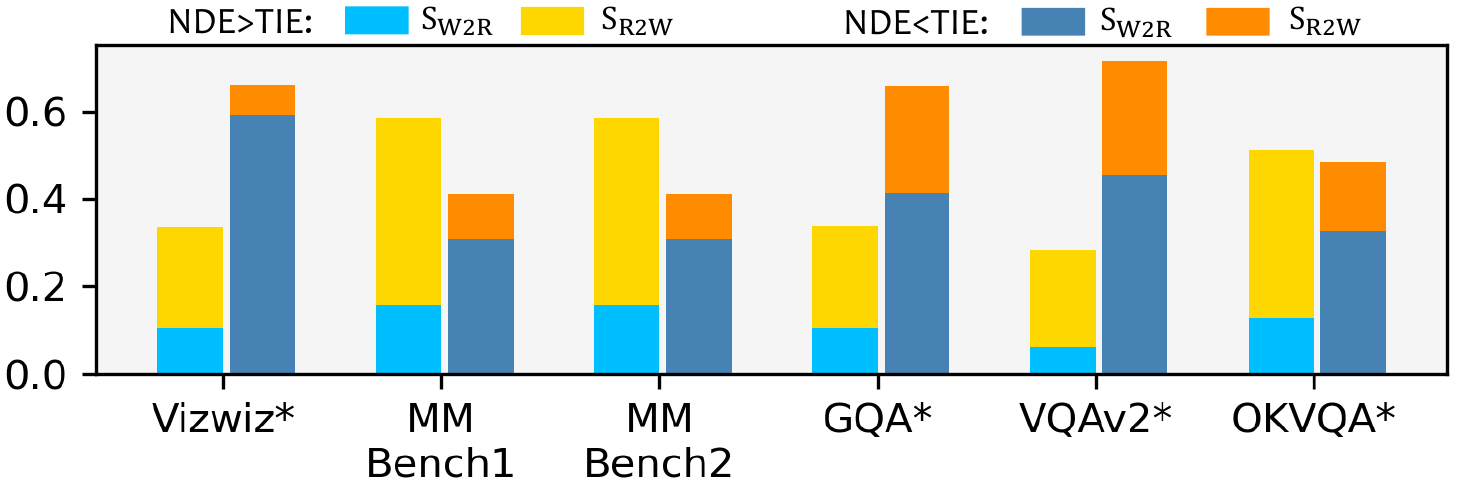}
\vspace{-1em}
    \caption{(1) MMBench1 and MMBench2 are two subtasks in MMBench: atribute\_reasoning and finegrained\_perception. (2) $\text{S}_{\text{W2R}}$ stands for samples on which CoG is helpful; $\text{S}_{\text{R2W}}$ means samples on which CoG is harmful. For samples with TIE\textgreater NDE, CoG is helpful for most of them.}
    \vspace{-1.em}
    \label{fig:compare}
\end{figure}

\subsubsection{Ablation of the Proposed Components}

% \begin{table*}[ht]
% \centering
% \footnotesize
% \begin{tabular}{l|ccccccccccc}
% \toprule
% LLaVA                  & \rotatebox{0}{MME} & \rotatebox{0}{\makecell[c]{SEED\\Bench}} & \rotatebox{0}{\makecell[c]{MM\\Bench}} & \rotatebox{0}{POPE}  & \rotatebox{0}{VSR} &\rotatebox{0}{\makecell[c]{Wino\\ground}*}&\rotatebox{0}{OKVQA*}&\rotatebox{0}{VQAv2*}&\rotatebox{0}{Vizwiz*}&\rotatebox{0}{GQA*}&\rotatebox{0}{CLEVR*}\\
% \midrule
% \midrule

% +Causal-CoG &&40.66&43.38&70.44&58.92&63.75&32.34&&41.53&& \\
% \midrule
% \quad w/o Causality Filter &&&39.64&70.21&56.87&62.50&29.56&&40.60 \\
% \midrule
% \quad w/o \scriptsize${Top\text{-}k}_{SIM^{c}}$ &&&42.89&&58.76&63.75&31.15&&39.44\\
% \quad \quad w/o \scriptsize$SIM^{c}$ &&&43.52&&58.92&66.25&32.74&&42.23\\
% \midrule
% \quad w/o \scriptsize${Top\text{-}k}_{TIE^{c}}$&&&43.37&&59.82&63.75&31.94&&40.84 \\
% \quad \quad w/o \scriptsize$TIE^{c}$ &&&42.73&&58.76&60.00&31.15&&38.28 \\
% \midrule
% \quad w/o \scriptsize${Top\text{-}k}_{SIM^{c},TIE^{c}}$&&&42.79&&58.67&62.50&31.35&&39.21 \\
% \quad \quad w/o \scriptsize$SIM^{c}$ &&&43.44&&59.17&65.00&32.74&&41.30\\
% \quad \quad w/o \scriptsize$TIE^{c}$ &&&42.74&&58.59&60.00&30.75&&37.82\\
% \midrule
% \midrule
% \makecell[l]{Majority Vote (unweighted)} \\
% +Causality Filter \\
% \bottomrule
% \end{tabular}
% \caption{}
% \label{tab:ablation}
% \end{table*}

\begin{table*}[ht]
\centering
\footnotesize
\begin{tabular}{l|cccccccccc}
\toprule
LLaVA                  & \rotatebox{0}{MME} & \rotatebox{0}{\makecell[c]{SEEDBench}} & \rotatebox{0}{\makecell[c]{MMBench}} & \rotatebox{0}{POPE}  & \rotatebox{0}{VSR} &\rotatebox{0}{\makecell[c]{Wino\\ground}*}&\rotatebox{0}{OKVQA*}&\rotatebox{0}{VQAv2*}&\rotatebox{0}{Vizwiz*}&\rotatebox{0}{GQA*}\\
\midrule
\midrule
Casusal-CoG &61.23&40.66&43.52&70.85&58.92&66.25&32.74&49.44&42.23&41.45 \\
\quad w/o \scriptsize$\text{Top-k}_{\text{TIE}^\text{c}}$ &61.58&39.93&43.44&70.66&59.17&65.00&32.74&49.49&41.30&40.89\\
\quad w/o Causality filter &58.43&37.60&41.62&69.07&58.02&60.00&29.17&44.26&38.98&38.58 \\

\midrule
\midrule
Weighted sum (likelihood)&59.94&39.89&42.69&70.03&58.10&60.00&30.56&46.18&37.59&39.46 \\
\quad w/o Causality filter &59.13&38.56&41.37&70.12&58.10&62.50&28.97&43.75&37.82&38.90 \\
\midrule
\midrule
Weighted sum (similarity) &60.23&40.23&42.74&70.09&58.59&60.00&30.75&47.99&37.82&38.82\\
\quad w/o Causality filter&59.34&38.93&41.62&70.20&58.02&62.50&29.37&47.20&38.52&38.11\\
\midrule
\midrule
Unweighted sum &60.24&40.17&42.71&70.03&58.10&60.00&30.75&47.90&38.05&38.75 \\
\quad w/o Causality filter &59.43&38.86&41.21&70.12&57.61&62.50&29.17&47.11&38.75&38.03\\

\bottomrule
\end{tabular}
\caption{Results of ablation and analysis experiments. (1) Either we remove the $\text{Top-}k$ aggregating strategy or the causality filter, performance drops on most datasets, signifying the necessity of these two modules. (2) During the aggregating stage, if the candidates' answers are aggregated based on other values, \emph{e.g.}, likelihood of each answer, similarity between context and image, or simple majority vote, we find that Causal-CoG does not perform as well as before, showing the importance of $\text{TIE}^{\text{c}}$ values.}
\vspace{-1.5em}
\label{tab:ablation}
\end{table*}

There are three main components in Causal-CoG: \textbf{Context Generation}, \textbf{Causality Filter} and \textbf{Top-$k$ Aggregation}. We conduct the ablation experiments of Causality Filter and Top-$k$ Aggregation respectively on 10 benchmarks mentioned above, illustrating the necessity and effectiveness of these two components.

\noindent
\textbf{Candidates with Higher $\text{TIE}^{\text{c}}$ are More Important.}
Table~\ref{tab:ablation} shows that in Causal-CoG, if we consider all candidates' $\text{TIE}^{\text{c}}$ when aggregating candidates' answers instead of using the top-$k$ strategy, accuracy on most benchmarks drops.

\begin{figure}[t]
    \centering
    \includegraphics[width=
\linewidth]{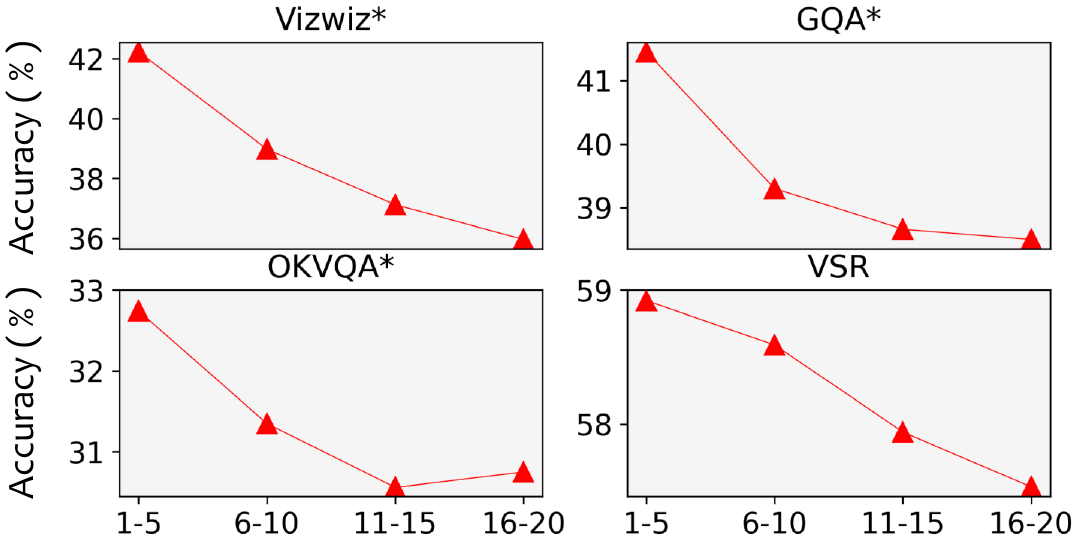}
    \caption{During aggregation, instead of aggregate candidates' answers based on top-k $\text{TIE}^{\text{c}}$ value, if we aggregate the answers from candidates with $6${th} to $10${th}, $11${th} to $15${th}, $16${th} to $20${th} high $\text{TIE}^{\text{c}}$ values, the performance drops a lot, showing that candidates with higher $\text{TIE}^{\text{c}}$ value are better when aggregating. }
    \vspace{-1.5em}
    \label{fig:topkacc}
\end{figure}

In Causal-CoG, when aggregating top-$k$ candidates, we select candidates with the $1${st} to $5${th} high $\text{TIE}^{\text{c}}$. We also explore the situation that uses candidates with $6${th} to $10${th}, $11${th} to $15${th}, $16${th} to $20${th} high $\text{TIE}^{\text{c}}$ to do weighted aggregation. As shown in Fig.~\ref{fig:topkacc}, using candidates with lower $\text{TIE}^{\text{c}}$ cause a significant accuracy drop, showing the effectiveness of $\text{TIE}^{\text{c}}$ as the aggregation weight.

\noindent
\paragraph{The Causality Filter Works by Identifying Samples on which CoG is Helpful.}
In Table~\ref{tab:ablation}, results without causality filter drop a lot compared with ones with causality filter, signifying the importance of causality filter in Causal-CoG.

The motivation of designing causality filter is to select the samples, where context has more consequence on the answer than image, \emph{i.e.}, TIE\textgreater NDE. Thus, we count the samples that can benefit from CoG, \emph{i.e.}, CoG corrects these samples' answers from wrong to right, termed as $\text{S}_{\text{W2R}}$, and samples that are harmed by CoG, \emph{i.e.}, the original answers output by MLM directly are right but CoG changes them to wrong, termed as $\text{S}_{\text{R2W}}$. The distribution of $\text{S}_{\text{W2R}}$ and $\text{S}_{\text{R2W}}$ in the samples set with NDE\textgreater TIE and NDE\textless TIE are illustrated as in Fig.~\ref{fig:compare}. As we can see, in samples with NDE\textless TIE,  $\text{S}_{\text{W2R}}$ accounts for the majority, \emph{i.e.}, CoG works well on most samples with NDE\textless TIE, thus we should apply CoG on these samples. While in samples with NDE\textgreater TIE,  $\text{S}_{\text{R2W}}$ is the most, \emph{i.e.}, CoG makes errors on most samples with NDE\textgreater TIE, so we should avoid applying CoG on these samples. These insightful results inform us how the causality filter works: \textbf{selecting the samples that can benefit from CoG}.
\vspace{-1em}

\subsubsection{Additional Analysis}
\noindent
\textbf{Other Aggregation Strategies.}
In the aggregation strategy we used in this paper, $\text{TIE}^{\text{c}}$ is used as the weights. However, there are other measures we can use as the weights when aggregating. For example, we can use each answer's likelihood or the similarity\footnote{The similarity is calculated using pretrained \textit{openai/clip-vit-patch14}} between the image and the generated context as the weights. Table~\ref{tab:ablation} shows the results using the answer's likelihood and the similarity between the given image and the generated context as weights to aggregate in Causal-CoG's framework. We also take place the weighted aggregation with unweighted aggregation (simple majority vote), and results are shown in Table~\ref{tab:ablation}.

We find that aggregating with $\text{TIE}^{\text{c}}$ as weights outperforms the other three aggregation strategies, informing us that $\text{TIE}^{\text{c}}$ is a better metric for aggregation. 

Besides, when removing the causal filter, performance drops even if candidates' answers are aggregated by the other three strategies, further showing the effectiveness of causality filter.

\noindent
\textbf{Contextual Information Assists VQA Tasks in MLM.}
To verify the direct effect of contextual information, we directly use the first generated candidate's answer as the final answer (getting rid of aggregation in Causal-CoG), termed as \textbf{Naive-CoG}, results are shown in Table~\ref{tab:overall}. Naive-CoG improves MLM's performance on 10 benchmarks, showing the helpfulness of contextual information itself for VQA.

\begin{figure}[t]
    \centering
    \includegraphics[width=
0.9\linewidth]{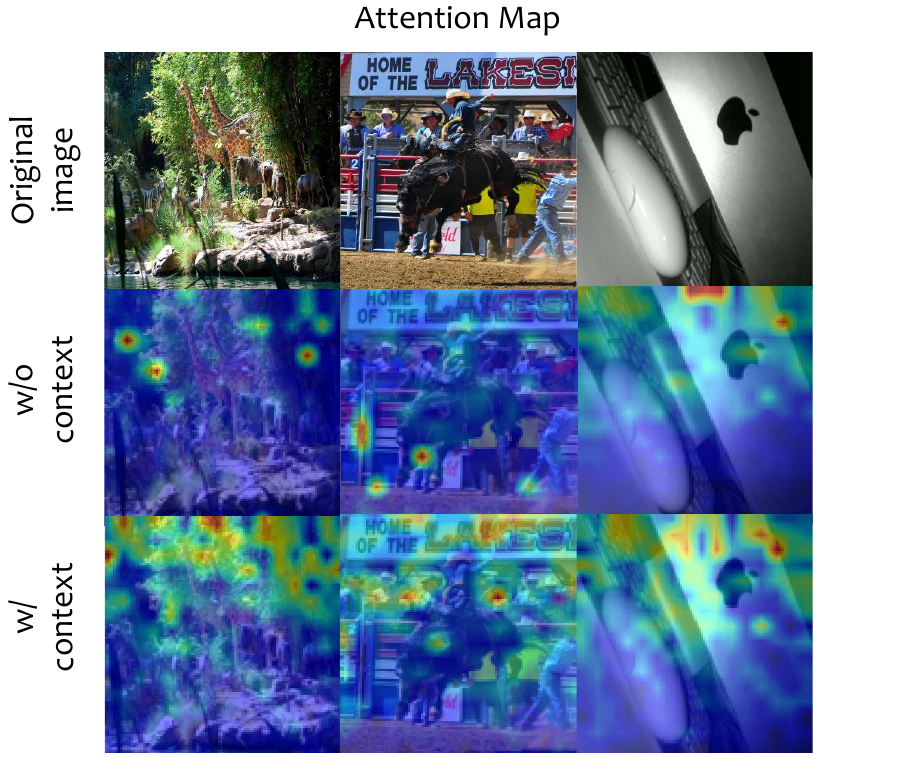}
    \caption{This figure reflects the changes in attention distribution after being provided context. (1) In these three samples, the question is: `` Is there a cow/horse/screen in this image?" And the attention map is queried by the right answer of this question above. (2) After providing context when doing VQA, MLM pays more attention to the context in the images, which is helpful for answering the question. For example, in the left-most image, provided with context, the trees and various animals are attended, so MLM can infer to the existence of a cow based on the contextual information and the image.}
    \vspace{-1.5em}
    \label{fig:attenmap}
\end{figure}

We also compare some samples' attention maps on the image queried by the answer of this visual question with and without Causal-CoG in Fig.~\ref{fig:attenmap}. As we can see, Causal-CoG helps the MLM to attend the context region of the image.

\noindent
\textbf{Causal-CoG on Other MLMs.}
Besides LLaVA and LLaVA-v1.5, we also apply Causal-CoG on MiniGPT-4~\cite{zhu23minigpt4}, and compare the results with ones without Causal-CoG on MME~\cite{fu2023mme}. Causal-CoG boosts MiniGPT-4's accuracy on MME's coarse-grained perception tasks by +2.09\%, showing the Causal-CoG can generalize to other MLMs. More results are provided in supplementary material.

 \section{Conclusion}
\vspace{-0.3em}
Motivated by contextual information can help MLM answer the visual questions better, we propose Causal-CoG to boost MLM's performance on VQA. In Causal-CoG, we also design a causality filter to determine whether the contextual information is helpful, thus deciding if we should use the answer from CoG or not. We prove the effectiveness of Causal-CoG by conducting the experiments over 10 VQA benchmarks. Causal-CoG improves the accuracy over all these 10 benchmarks significantly compared to the original MLM.

%%%%%%%%% REFERENCES
\clearpage
\newpage
{
 %\begin {thebibliography}
  
   \small
   \bibliographystyle{ieeenat_fullname}
    \bibliography{main}
    %\end{thebibliography}
}

\newpage
\appendix
\clearpage
\setcounter{page}{1}
\onecolumn
\maketitlesupplementary
\section{Ablation of Number of Candidates}
To explore how the number of candidates can affect Causal-CoG's performance, we conduct a series of experiments with different numbers of candidates, and see how the result changes. In Table~\ref{tab:candidates}, we apply Causal-CoG with 5, 10, 15 and 20 candidates respectively on LLaVA~\cite{liu23llava}. In Causal-CoG, we use candidates to estimate the $\text{TIE}$ and $\text{NDE}$ values and the final answer is the aggregating result of candidates. More candidates lead to more accurate estimation and aggregating results.

\begin{table}[h]
    \centering
    \footnotesize
    \begin{tabular}{l|ccccc}
    \toprule
    \makecell{Number of\\ Candidates} & VQAv2*&GQA*&OKVQA*&Vizwiz*&VSR \\
    \midrule
    5 &49.30&40.33&31.75&40.84&\textbf{59.33} \\
    \midrule
    10 &49.07&40.65&31.94&\textbf{42.00}&59.17 \\
    \midrule
    15 &\textbf{49.63}&40.89&31.94&41.53&58.43\\
    \midrule
    20 &49.44&\textbf{41.45}&\textbf{32.74}&41.30&58.59\\
    \bottomrule
    \end{tabular}
    \caption{Accuracy with different numbers of candidates.}
    \label{tab:candidates}
\end{table}

\section{Ablation of $k$'s Value in Top-$k$ Aggregation}
We have proved the importance of $\text{TIE}^{\text{c}}$'s value in the experiment section during aggregation. In our paper, we set $k$=5 on most benchmarks when doing aggregation. Here we conduct the ablation analysis of $k$'s value to see how it affects the Causal-CoG's performance. In Table~\ref{tab:k}, we apply Causal-CoG with $k$ whose range is \{1, 5, 10, 15, 20\} on LLaVA~\cite{liu23llava}.

\begin{table}[h]
    \centering
    \footnotesize
    \begin{tabular}{l|ccccc}
    \toprule
    $k$'s value&VQAv2*&GQA*&OKVQA*&Vizwiz*&VSR \\
    \midrule
    1 &49.25&\textbf{41.53}&\textbf{32.94}&\textbf{42.92}&59.08\\
    \midrule
    5 &49.44&41.45&32.74&42.23&58.92\\
    \midrule
    10 &49.39&41.37&32.74&41.53&58.18\\
    \midrule
    15 &49.35&41.05&32.74&41.53&\textbf{59.33}\\
    \midrule
    20 &\textbf{49.49}&40.89&32.74&41.30&59.17\\
    \bottomrule
    \end{tabular}
    \caption{Accuracy results with different $k$'s value.}
    \label{tab:k}
\end{table}

\section{Combine Multiple Metrics When Aggregating}
In Causal-CoG, we aggregate the candidates' answers with $\text{TIE}^{\text{c}}$ as weights, and we also explore the performance of using other metrics as weights to aggregate answers, \emph{e.g.}, the similarity between context and image, and the likelihood of the answer. When applying Causal-CoG, the context may be inaccurate because of the limited ability of MLM. So we try to consider the quality of the generated context when doing aggregation. Thus, in this section, we combine $\text{TIE}^{\text{c}}$ and similarity, termed as $\text{SIM}^{\text{c}}$, to aggregate the candidates' answers. 

We consider $\text{TIE}^{\text{c}}$ and $\text{SIM}^{\text{c}}$ during the aggregation stage, instantiated as using the sum of top-$k$ $\text{TIE}^{\text{c}}$ and $\text{SIM}^{\text{c}}$ as weights. In Table~\ref{tab:sim}, Causal-CoG with $\text{TIE}^{\text{c}}$ and $\text{SIM}^{\text{c}}$ as aggregation metric can harm the performance, which we think is caused by the limitation of the $\text{SIM}^{\text{c}}$ calculating methods. The $\text{SIM}^{\text{c}}$ is calculated by pretrained CLIP~\cite{radford21clip} from OpenAI. For CLIP, the length of the text encoder is 77, which is limited for most context generated by MLM, \emph{i.e.}, most context's length is more than 77 and we need to truncate these contexts to calculate the similarity, thus the $\text{SIM}^{\text{c}}$ values could be inaccurate.

\begin{figure}[h]
    \centering
\includegraphics[width=0.7\linewidth]{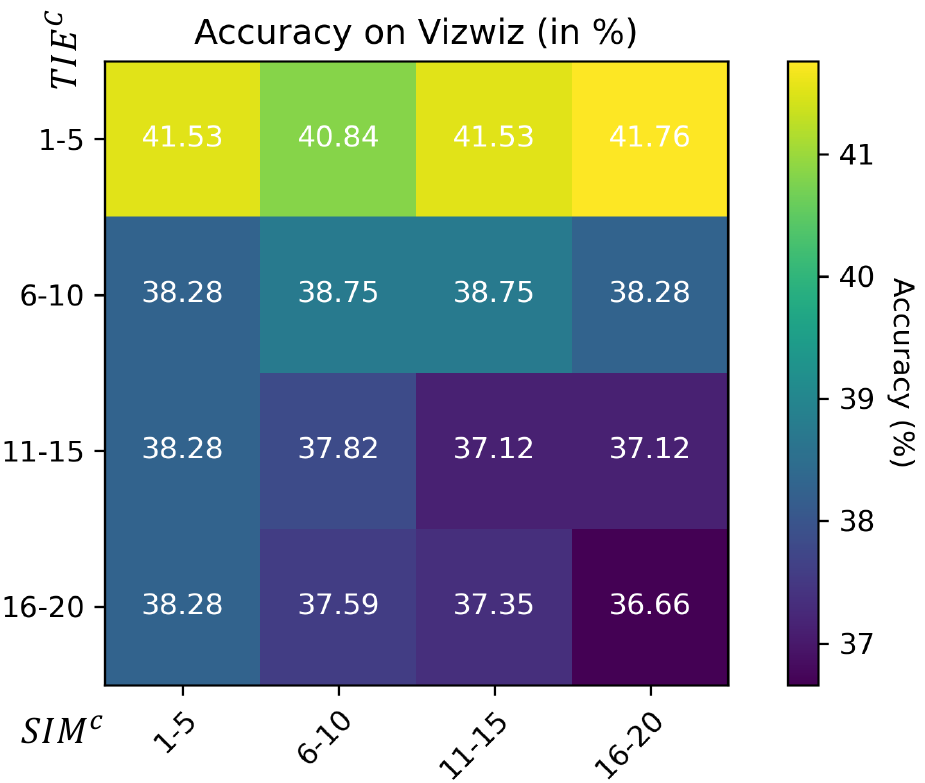}
    \caption{Accuracy on Vizwiz* with $\text{SIM}^{\text{c}}$ and $\text{TIE}^{\text{c}}$ in different intervals.}
    % \vspace{-1.0em}
    \label{fig:accvizwiz}

\end{figure}

We also explore the top-$k$ operation's consequence in this two-metric aggregation procedure. As shown in Fig.~\ref{fig:accvizwiz}, aggregating with the last 5 high $\text{SIM}^{\text{c}}$ and $\text{TIE}^{\text{c}}$ performs poorly on Vizwiz*, which shows that $\text{SIM}^{\text{c}}$ and $\text{TIE}^{\text{c}}$ can signify the low-quality candidates.

\begin{table}[h]
    \centering
    \footnotesize
    \begin{tabular}{c|c}
    \toprule
    Aggregation metric&Accuracy on Vizwiz* \\
    \midrule
    $\text{TIE}^{\text{c}}$ &\textbf{42.23}\\
    \midrule
    $\text{TIE}^{\text{c}}$ and  $\text{SIM}^{\text{c}}$&41.53\\
    \bottomrule
    \end{tabular}
    \caption{Comparison of accuracy on Vizwiz* w/ and w/o $\text{SIM}^{\text{c}}$. }
    \label{tab:sim}
    
\end{table}

\section{Ablation of Sampling Methods During Generation}
We instantiate the context generation by top-$k$\footnote{This top-$k$ is totally different from Top-$k$ aggregation strategy in our proposed Causal-CoG, \emph{i.e.}, this top-$k$ is a sampling method which is widely used in language models.}~\cite{fan18topk} sampling strategy, where $k=40$, temprature $t=0.9$. We can control the diversity of the generated contexts by setting $k$ and $t$ to different values. Here, we conduct experiments with different $t$' values, shown in Table~\ref{tab:sampling}. We can see that, setting $t$ to 0.9 or 0.7 leads to better performance, \emph{i.e.}, diverse contexts bring benefit to Causal-CoG.

\begin{table}[h]
\centering
\footnotesize
\begin{tabular}{l|ccccc}
\toprule
 \makecell{Sampling\\Setting}& VQAv2* & GQA* & OKVQA* & Vizwiz* & VSR \\
 \midrule
 t=0.9,k=40& 49.44 & \textbf{41.45} & \textbf{32.74} & 42.23 & 58.92 \\
 \midrule
 t=0.7,k=40& \textbf{52.28} & 41.21 & 32.34 & \textbf{42.69} & 59.57 \\
 \midrule
 t=0.5,k=40& 50.70 & 40.02 & 32.54 & 41.76 & \textbf{60.80} \\
 \midrule
 t=0.3,k=40 &50.75&40.33&32.74&41.30&60.64\\
 \midrule
 t=0.1,k=40 &49.67&39.94&32.74&41.53&60.80 \\
 \bottomrule
\end{tabular}
\caption{Accuracy results with different temperatures.}
\label{tab:sampling}
\end{table}

\section{Statistics of Benchmarks}
In Table~\ref{tab:staticdataset}, the statistics of each benchmark, including version and number of samples, are listed.
\begin{table}[h]
    \centering
    \footnotesize
    \begin{tabular}{l|c|c}
    \toprule
    Benchmark & Version & Number of Samples \\
    \midrule
    MME&-&1974 \\
    SEEDBench&-&14233 \\
    MMBench&dev&4377 \\
    POPE&Popular,Random,Adversarial&8910 \\
    VSR&-&1222\\
    Winoground&ReForm-Eval&60 \\
    OKVQA&ReForm-Eval&504 \\
    VQAv2&ReForm-Eval&2144 \\
    Vizwiz&ReForm-Eval&431 \\
    GQA&ReForm-Eval&1257 \\
    \bottomrule
    \end{tabular}
    \caption{Statistics of 
 each benchmark.}
    \label{tab:staticdataset}
\end{table}

\section{Full List of System Prompts Used in Ensemble Method}
In the \textit{Ensemble} method, we use 5 different system prompts to generated 5 answers and then ensemble these answers by majority vote. Full list of the system prompts is shown in Table~\ref{tab:systemprompt}.

\begin{table*}[h]
\centering
\begin{tabularx}{\textwidth}{X}
\toprule
System Prompt \\
\midrule
A chat between a curious user and an artificial intelligence assistant. The assistant is able to understand the visual content that the user provides, and assist the user with a variety of tasks using natural language.
 \\
 \midrule
 You are a helpful language and vision assistant. You are able to understand the visual content that the user provides, and assist the user with a variety of tasks using natural language.
 \\
 \midrule
 You are a helpful, respectful and honest assistant. Always answer as helpfully as possible, while being safe.  Your answers should not include any harmful, unethical, racist, sexist, toxic, dangerous, or illegal content. Please ensure that your responses are socially unbiased and positive in nature.
 \\
 \midrule
 Give the following image. You will be able to see the image once I provide it to you. Please answer my questions.
 \\
 \midrule
 A chat between a curious human and an artificial intelligence assistant. The assistant gives helpful, detailed, and polite answers to the human's questions. \\
 \bottomrule

\end{tabularx}
\caption{System prompts used in \textit{Ensemble} method.}
\label{tab:systemprompt}
\end{table*}

\section{One-shot Sample Used in One-shot Method}

In the \textit{One-shot} method, the in-context sample we used is shown in Table~\ref{tab:oneshot}.

\begin{table*}[h]
\centering
\begin{tabular}{ccc}
\toprule
 Image& Question &  Answer\\
 \midrule
 \raisebox{-0.5\height}{\includegraphics[width=0.2\textwidth]{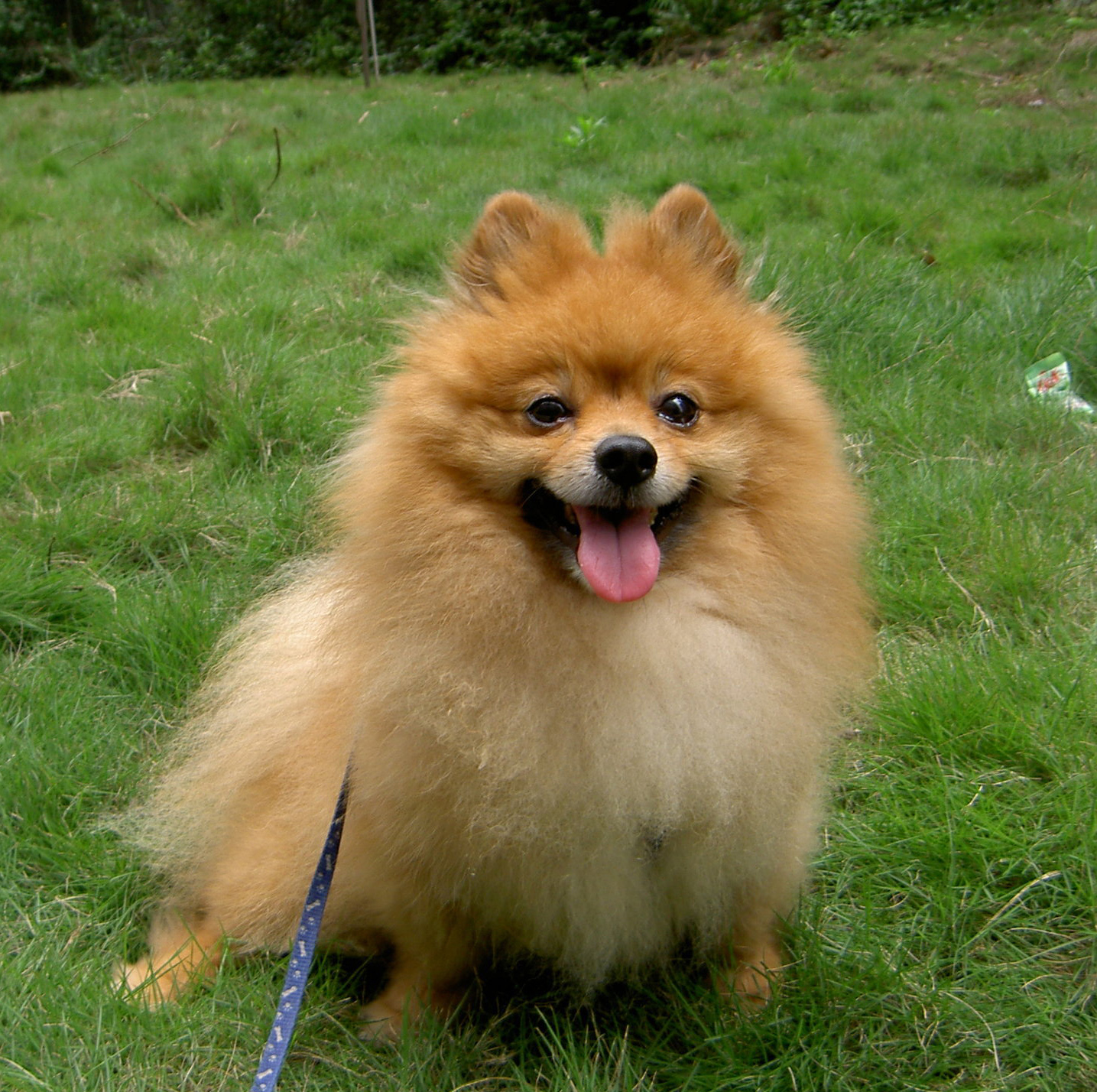}}& What is the animal in this image? & There is a dog in this image. \\
 \bottomrule
\end{tabular}
\caption{Sample used in \textit{One-shot} method.}
\label{tab:oneshot}
\end{table*}

\section{Causal-CoG on Other MLMs}

We apply Causal-CoG on MiniGPT-4, the results are listed in Table~\ref{tab:minigpt}. 
\begin{table*}[h]
\centering
\begin{tabular}{lccccc|cc}
\toprule
&\multicolumn{5}{c}{MME}&\multicolumn{2}{c}{MMBench} \\
&Existence&Color&Position&Code&Commonsese&Relation-reasoning&Coarse-perception\\
\midrule
 MiniGPT-4& 61.67&46.67&55.00&47.50&50.00&23.37&53.44   \\
 \midrule
 +Causal-CoG& 63.33&50.00&58.33&52.50&53.57 &24.27&53.99  \\
 \midrule
$\Delta$ &+1.66&+3.33&+3.33&+5.00&+3.57&+0.90&+0.55    \\
 \bottomrule
\end{tabular}
\caption{Accuracy results of applying Causal-CoG on MiniGPT-4. We select some subtasks from MME and MMBench respectively, some of them are cognition tasks and the other are perception tasks.}
\label{tab:minigpt}
\end{table*}

\section{Task Split of Cognition and Perception in MME, SEEDBench and MMBench}
In MME~\cite{fu2023mme}, SEEDBench~\cite{Li23seedbench} and MMBench~\cite{Liu23mmbench}, subtasks are split into two groups: cognition and perception, as shown in Table~\ref{tab:tasksplit}.
\begin{table*}
    \centering
    \begin{tabular}[h]{l|c|c}
    \toprule
    &Cognition&Perception \\
    \midrule
    MME& \makecell{Code Reasoning; Text Translation;\\ Numerical Calculation; Commonsense Reasoning}&\makecell{Existence; Color; Count; Position; OCR;\\ Scene; Celebrity; Artwork; Posters} \\
    \midrule
    SEEDBench  & \makecell{Spatial Relation; Instance Interaction;\\ Visual Reasoning; Text Recognition} &\makecell{Scene Understanding; Instance Identity; Instance\\ Attribute; Instance Location; Instance Counting} 
 \\
    \midrule
    MMBench &\makecell{Attribute Reasoning; Logic \\Reasoning; Relation Reasoning}&\makecell{Finegrained Perception (single-instance); Finegrained \\Perception (cross-instance); Coarse Perception} \\
    \bottomrule
    \end{tabular}
    \caption{Task split of MME, SEEDBench and MMBench.}
    \label{tab:tasksplit}
\end{table*}

\section{More Examples where Context Helps Doing VQA}

In this section, we apply CoG on LLaVA-v1.5-13B~\cite{liu23llava15} and QwenVL-7B~\cite{bai23qwenvl} with an open-ended form. Qualitative results are shown in Tables~\ref{tab:r1},\ref{tab:r2},\ref{tab:r3}.

\begin{table*}[h]
\centering
\begin{tabularx}{\textwidth}{>{\hsize=.5\hsize\linewidth=\hsize}c|>{\hsize=.5\hsize\linewidth=\hsize}X|>{\hsize=.5\hsize\linewidth=\hsize}X|>{\hsize=1.5\hsize\linewidth=\hsize}X|>{\hsize=.5\hsize\linewidth=\hsize}X}
\toprule
 Image&  Question & Original Answer& Context &  Answer with Context\\
 \midrule
\raisebox{-1.5\height}{\includegraphics[width=0.2\textwidth]{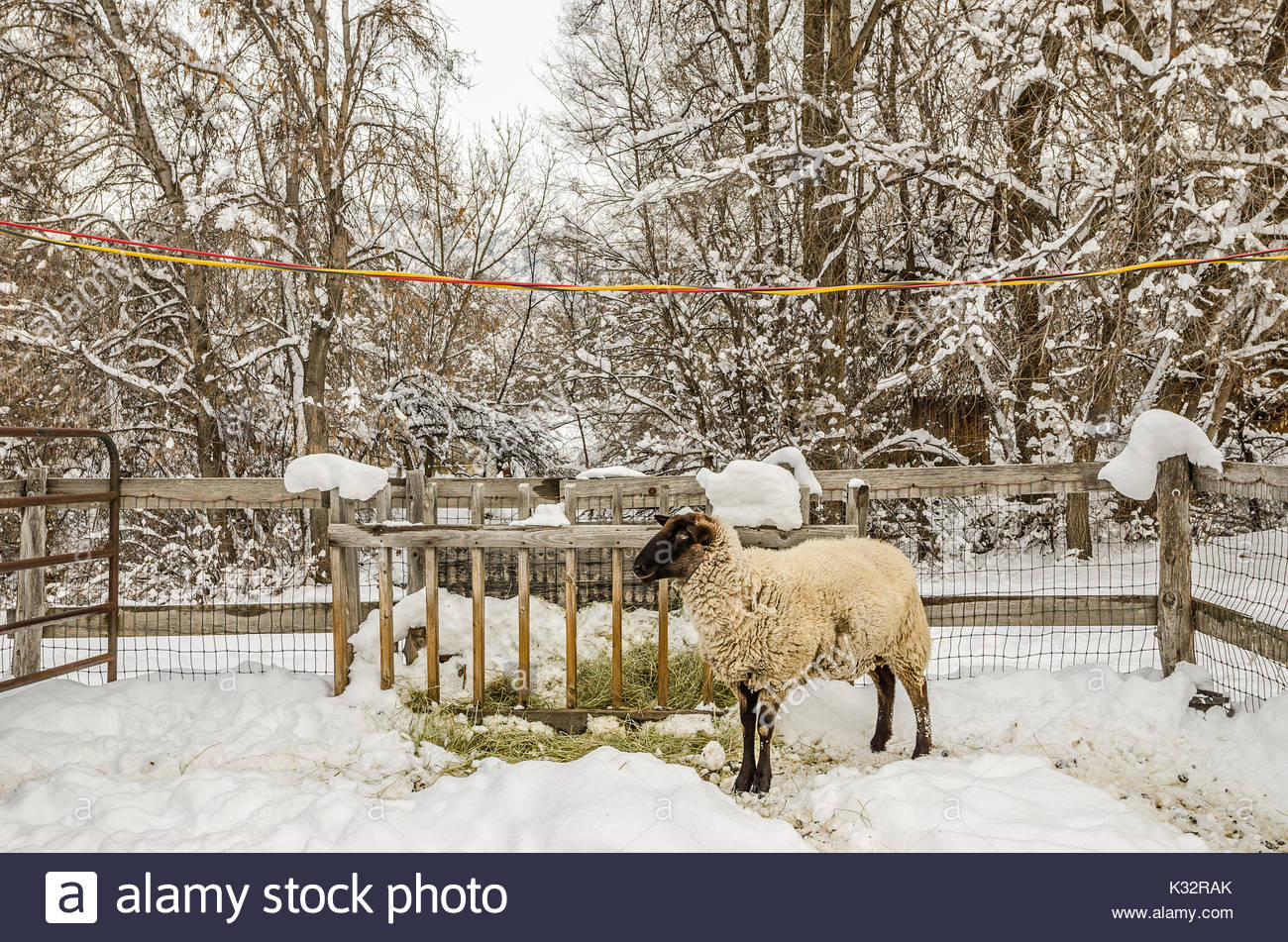}}& There are two sheeps in this image, am I right? &Yes, there are two sheeps in the image.& The image features a snowy scene with a sheep standing in a fenced area. The sheep is positioned in the middle of the enclosure, surrounded by a snow-covered ground. The fence is made of wood and wire, providing a secure area for the sheep. &  No, there is only one sheep in the image. The other sheep you might be referring to is a different image or a part of the same image that is not visible in the current frame.\\
\midrule
 \raisebox{-1.2\height}{\includegraphics[width=0.2\textwidth]{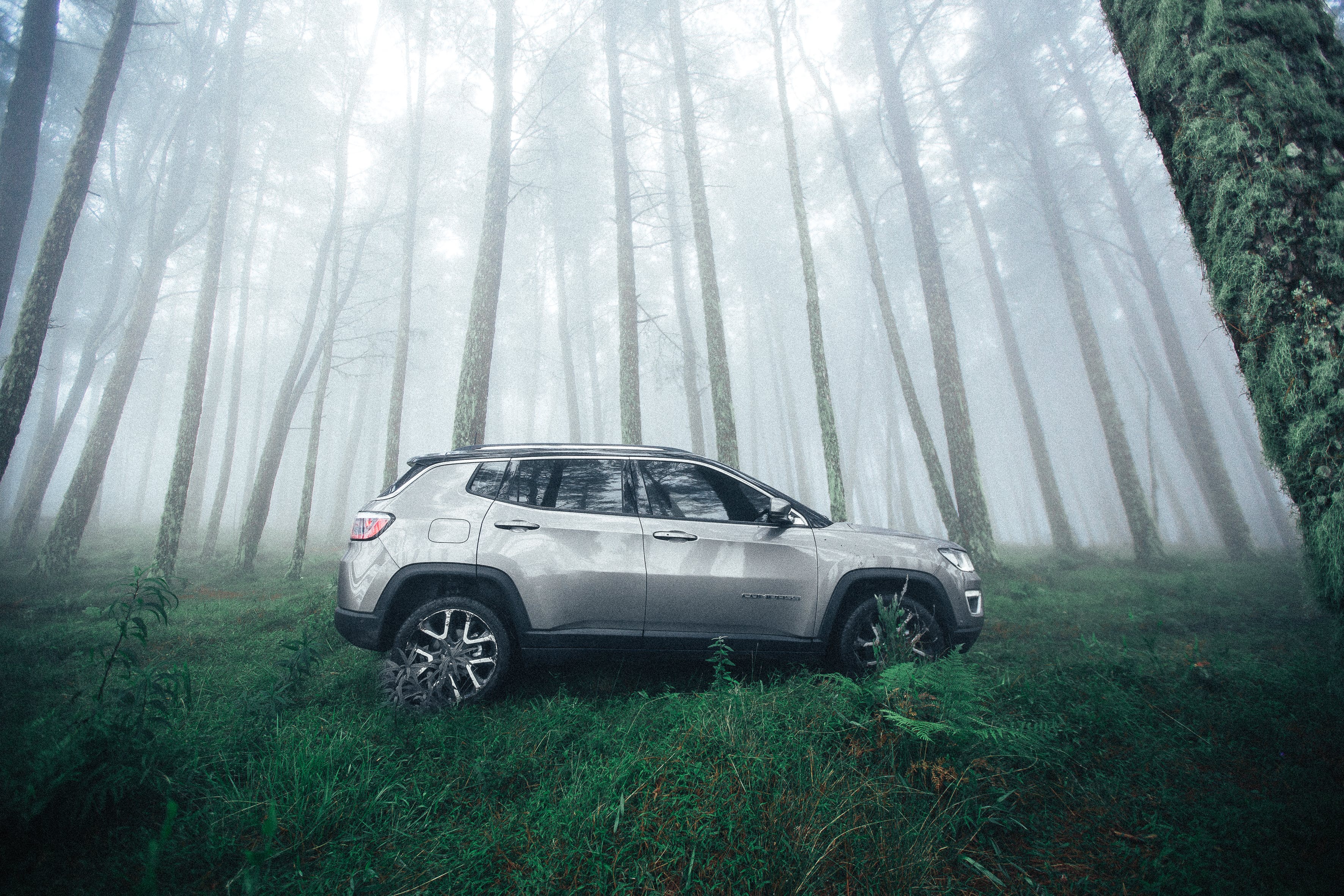}}& What is the color of this car? &The color of the car in the image is white.  & The image features a silver SUV parked in a forest, surrounded by tall trees. The vehicle is positioned on a grassy area, with its front end facing the viewer. The scene appears to be foggy, adding a sense of mystery to the setting. The SUV is the main focus of the image, with the trees in the background creating a serene and natural atmosphere.&The color of the car in the image is silver. \\
 \midrule
 \raisebox{-0.9\height}{\includegraphics[width=0.2\textwidth]{samples_cog/3.png}}&What is the spacial relationship between the phone and the cup?&The phone is located next to the cup on the table.&The image features a wooden desk with a variety of items on it. There is a cell phone placed towards the center of the desk, and a cup, possibly containing coffee or tea, is positioned slightly to the left of the phone. A bottle can be seen on the left side of the desk, and a green apple is located near the left edge of the desk.&The phone is positioned in front of the cup, with the cup being slightly to the left of the phone.\\
  \bottomrule
\end{tabularx}
\caption{Qualitative results of applying CoG on LLaVA-v1.5-13B.}
\label{tab:r1}
\end{table*}

\begin{table*}[h]
\centering
\begin{tabularx}{\textwidth}{>{\hsize=.5\hsize\linewidth=\hsize}c|>{\hsize=.5\hsize\linewidth=\hsize}X|>{\hsize=.5\hsize\linewidth=\hsize}X|>{\hsize=1.5\hsize\linewidth=\hsize}X|>{\hsize=.5\hsize\linewidth=\hsize}X}
\toprule
 Image&  Question & Original Answer& Context &  Answer with Context\\
 \midrule
  \raisebox{-1.5\height}{\includegraphics[width=0.2\textwidth]{samples_cog/4.png}}&Are there any kids in this image?&Yes, there are kids in the image. Some kids are walking down the nearby sidewalk.&The image features a silver fire hydrant located on a city sidewalk near a street. The fire hydrant is positioned next to a metal pole, and it appears to be rusty. There are several cars parked on the street, including one close to the fire hydrant, and others further down the road. In the scene, there are also a couple of people, one near the center of the image and another closer to the right side. A stop sign is visible on the right side of the street, indicating a traffic control point in the area.&No, there are no kids in the image. The presence of a person in the scene could be due to a pedestrian or someone walking their dog, but it is not clear if they are a child or an adult.\\
\midrule
\raisebox{-2\height}{\includegraphics[width=0.2\textwidth]{samples_cog/6.png}}& Is it rainy in this picture?&Yes, it is raining in the picture, as evidenced by the man holding an umbrella while riding his bicycle.&The image features a man riding a bicycle down a street, with an umbrella attached to the back of his bike. He is shirtless and appears to be enjoying his ride. There are several other people in the scene, some of them standing or walking along the street. In addition to the man on the bicycle, there are two motorcycles parked on the side of the street, and a car is visible further down the road. The street is lined with various buildings, creating a lively urban atmosphere.&Although the man has an umbrella attached to his bike, the image does not explicitly show that it is raining. However, the presence of the umbrella suggests that there might be a chance of rain, or the man might be using it for protection from the sun.\\
 \bottomrule
\end{tabularx}
\caption{Qualitative results of applying CoG on LLaVA-v1.5-13B.}
\label{tab:r2}
\end{table*}

\begin{table*}[h]
\centering
\begin{tabularx}{\textwidth}{>{\hsize=.5\hsize\linewidth=\hsize}c|>{\hsize=.5\hsize\linewidth=\hsize}X|>{\hsize=.5\hsize\linewidth=\hsize}X|>{\hsize=1.5\hsize\linewidth=\hsize}X|>{\hsize=.5\hsize\linewidth=\hsize}X}
\toprule
 Image&  Question & Original Answer& Context &  Answer with Context\\
 \midrule
  \raisebox{-1.4\height}{\includegraphics[width=0.2\textwidth]{samples_cog/2.jpg}}&The color of this car is white, am I right?&Yes, the car in the image is white.&The image shows a silver Jeep Compass parked in a forest covered in fog. The fog creates a misty, ethereal atmosphere, while the Jeep is parked on a grass-covered hillside surrounded by trees. The tall trees on either side of the Jeep create a frame for the vehicle, emphasizing its position in the center of the image. The foggy forest setting and theJeep create a sense of adventure and exploration, as if the driver has taken a wrong turn and ended up in this magical, misty forest.&No, the car in the image is silver, not white.\\
\midrule
\raisebox{-1.3\height}{\includegraphics[width=0.2\textwidth]{samples_cog/8.jpg}}& The color of the hydrants is yellow, am I right?&The color of the hydrants is indeed yellow.&The image features a row of four tall, orange columns lined up on a sidewalk in a large city. These columns have a golden, circular design element on the middle of each, making them appear decorative. The columns are situated in front of a large, old building with gothic architecture, and behind these columns, there is another smaller building with a fence surrounding it. In the distance, on the left side of the image, there is a group of people, likely pedestrians enjoying the city. The scene has a frosty feel to it, suggesting that it was taken during the winter season.&The hydrants are described as old and rusty in the image description, so the color yellow would not be accurate. The hydrants are actually described as brown and rusty.\\
 \bottomrule
\end{tabularx}
\caption{Qualitative results of applying CoG on QwenVL-7B.}
\label{tab:r3}
\end{table*}

\end{document}